\documentclass{ieeeaccess}
\usepackage{cite}
\usepackage{amsmath,amssymb,amsfonts}
\usepackage{algorithmic}
\usepackage{graphicx}
\usepackage{textcomp}

\usepackage{booktabs}
\usepackage{multirow}
\usepackage{subfig}

\usepackage{soul}
\renewcommand\hl[1]{#1}

\begin{document}

\history{Date of publication xxxx 00, 0000, date of current version xxxx 00, 0000.}
\doi{10.1109/ACCESS.2023.0322000}

\title{Artifact Reduction in 3D and 4D \hl{Cone-beam Computed Tomography} Images with Deep Learning - \hl{A Review}}

\author{\uppercase{Mohammadreza Amirian}\authorrefmark{1},
Daniel Barco\authorrefmark{1}, 
Ivo Herzig\authorrefmark{2},
and Frank-Peter Schilling\authorrefmark{1}
}

\address[1]{Centre for AI (CAI), Zurich University of Applied Sciences (ZHAW), Winterthur, Switzerland (e-mail: mohammadreza.amirian@gmail.com,\{baoc,scik\}@zhaw.ch)}

\address[2]{Institute of Applied Mathematics and Physics (IAMP), Zurich University of Applied Sciences (ZHAW), Winterthur, Switzerland (e-mail: hezi@zhaw.ch)}

\tfootnote{This work was supported in part by Innosuisse under Grant 56768.1 IP-LS. }

\markboth
{Amirian \headeretal: Artifact Reduction in 3D and 4D Cone-beam Computed Tomography Images with Deep Learning - A Review}
{Amirian \headeretal: Artifact Reduction in 3D and 4D Cone-beam Computed Tomography Images with Deep Learning - A Review}

\corresp{Corresponding author: Frank-Peter Schilling (e-mail: scik@zhaw.ch).}

\begin{abstract}
Deep learning based \hl{approaches} have been used to improve image quality in cone-beam computed tomography (CBCT), a medical imaging technique often used in applications such as image-guided radiation therapy, implant dentistry or orthopaedics. In particular, while deep learning methods have been applied to reduce various types of CBCT image artifacts arising from motion, metal objects, or low-dose acquisition, a comprehensive review summarizing the successes and shortcomings of these approaches\hl{, with a primary focus on the type of artifacts rather than the architecture of neural networks,} is lacking in the literature. In this review,\hl{ the data generation and simulation pipelines, and artifact reduction techniques are specifically investigated for each type of artifact}. We provide an overview of deep learning techniques that have successfully been shown to reduce artifacts in 3D, as well as in time-resolved (4D) CBCT through the use of projection- and/or volume-domain optimizations, or by introducing neural networks directly within the CBCT reconstruction algorithms. 
\hl{Research gaps are identified to suggest avenues for future exploration. One of the key findings of this work is an observed trend towards the use of generative models including GANs and score-based or diffusion models, accompanied with the need for more diverse and open training datasets and simulations.}

\end{abstract}


\titlepgskip=-21pt

\maketitle

%%%%%%%%%%%%%%%%%%%%%%%%%%%%%%%%%%%%%%%%%%%%%%%%%%%%%%%%%%%%%%%%%%%%%%%%%%%%%%%%%%%%%%%%%%%%%%%%%%%%%%

\begin{figure*}[bth!]
	\centering
	\includegraphics[width=\linewidth]{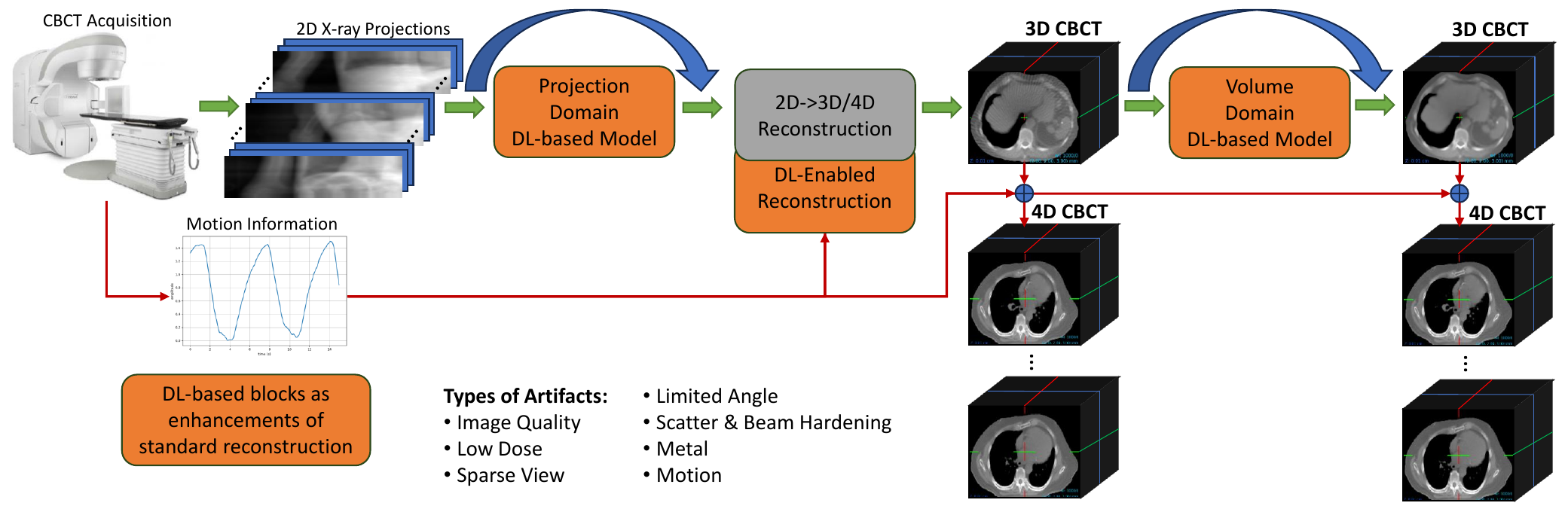}
	\caption{Visual Abstract: 
		An illustration of the CBCT acquisition process in IGRT for lung CBCT and the application of deep learning for artifact correction.
		The diagram depicts the acquisition of 2D projections (initial corrections such as scatter corrections have already been applied), including (optionally) time- and motion-related information (e.g. breathing amplitude signal), standard CBCT reconstruction (typically 2D$\rightarrow$3D), and DL-based components for image enhancement.
		Incorporating acquired temporal and motion information provides the opportunity to apply a projection binning which can be used to reconstruct 4D CBCT images (3D images at various states of motion).
		During the course of CBCT reconstruction, several types of artifacts (e.g. arising from cone-beam geometry, low dose, sparse view or limited angle scans, scatter, metal or beam hardening) can be mitigated through DL-based optimization in the projection and/or volume domain, or by improving (parts of) the reconstruction algorithm itself using neural networks.
		The illustration of a commerical radiotherapy system is adapted from~\cite{Shende2016CommissioningOT}.}
	\label{fig:visual_abstract}
\end{figure*}

\section{Introduction}
Cone-beam computed tomography (CBCT) is an imaging technique to acquire volumetric scans in medical domains such as implant dentistry, orthopaedics, or image-guided radiation therapy (IGRT). In particular, in the case of IGRT, on-board imaging mounted directly on radiotherapy machines is used to assess a patient’s current anatomy before radiation treatment sessions. Changes in anatomy during the treatment period and since the acquisition of the planning CT (pCT) can lead to inefficiencies in the treatment process.
Recent research has demonstrated that utilizing 3D or 4D \hl{(volumetric data with additional time dimension to track motion)} CBCT scans in IGRT~\cite{jaffray2002flat} improves patient positioning and dose calculation for radiotherapy sessions.

The quality of CBCT scans suffers from similar types of artifacts as for spiral/helical CT scans, including those arising from beam hardening and scatter effects, metal implants, and patient motion. In addition, new artifacts arise due to the cone-beam geometry.
Further, minimizing the radiation dose in radiotherapy is important for the safety of the patients. 
However, reducing the imaging dose per scan, acquiring fewer X-ray projections, or acquiring projection data from a limited angle can result in streak artifacts.

This paper provides an overview of the current body of research on artifact reduction in 3D and 4D CBCT with applications including, but not limited to, IGRT, aiming to improve scan quality while also minimizing the imaging radiation dose.
\hl{The significant variation in the methods and techniques used to mitigate different types of artifacts suggests to organize the literature based on the type of artifact. For instance, sparse-view artifacts can be addressed in the projection domain by interpolating new projections, but refining the original projections is not beneficial; however, motion artifact mitigation is possible through projection refinement. Further, the survey aims to present a clear picture of all necessary steps in the artifact mitigation process for all relevant types of artifacts individually.
}

In particular, we review the current state-of-the-art research which uses deep learning (DL)~\cite{Goodfellow-et-al-2016} to reduce various artifacts in CBCT scans, and we categorize the research based on the types of artifacts they address.
While Ref.~\cite{paysan_annpr2020} focuses broadly on the use of DL methods in IGRT, the closest literature reviews to our work are presented in references~\cite{spadea2021deep, manual_zhang_review_2022, manual_rusanov_review_2022}.
The first survey~\cite{spadea2021deep} is focused on synthetic CT generation from various types of input scans, including CBCT, with the aim to enhance the scan quality. 
Its content partially overlaps with what we present in Section~\ref{sec:image quality}.
However, it does not cover all the other artifacts which can degrade CBCT image quality as discussed after Section~\ref{sec:image quality}.
Ref.~\cite{manual_zhang_review_2022} discusses supervised, self-supervised, and unsupervised techniques for artifact reduction in CT scans, and it covers unrolling the reconstruction, as well as optimization methods in both the projection (raw 2D X-ray images) and volume (reconstructed 3D images) domains.
However, it is essential to note that Ref.~\cite{manual_zhang_review_2022} primarily focuses on CT scans, which differs from the main focus of this work, namely CBCT scans.
The third survey~\cite{manual_rusanov_review_2022}  provides an in-depth literature analysis, considering criteria such as anatomy, loss functions, model architectures, and training methods for supervised learning specifically applied to CBCT scans. In our work, instead of dividing the literature based on the deep learning methods, we group the research based on the type of artifacts, discussing results employing projection- and/or volume-domain optimization, dividing the methods based on the type of supervision, and also including research addressing time-resolved 4D CBCT reconstruction. 

Artifacts in CBCT images can principally be reduced by optimizations in the projection, volume, or dual-domain (both projections and volumes), as well as by DL-enabled reconstruction.
This survey presents an overview of deep learning techniques able to reduce artifacts in 3D as well as time-resolved 4D CBCT using optimizations in the above domains, and through novel CBCT reconstruction methods.
Furthermore, it addresses the challenges and limitations associated with these approaches and provides recommendations for future research directions.

\begin{figure}
    \centering
    \includegraphics[width=\linewidth]{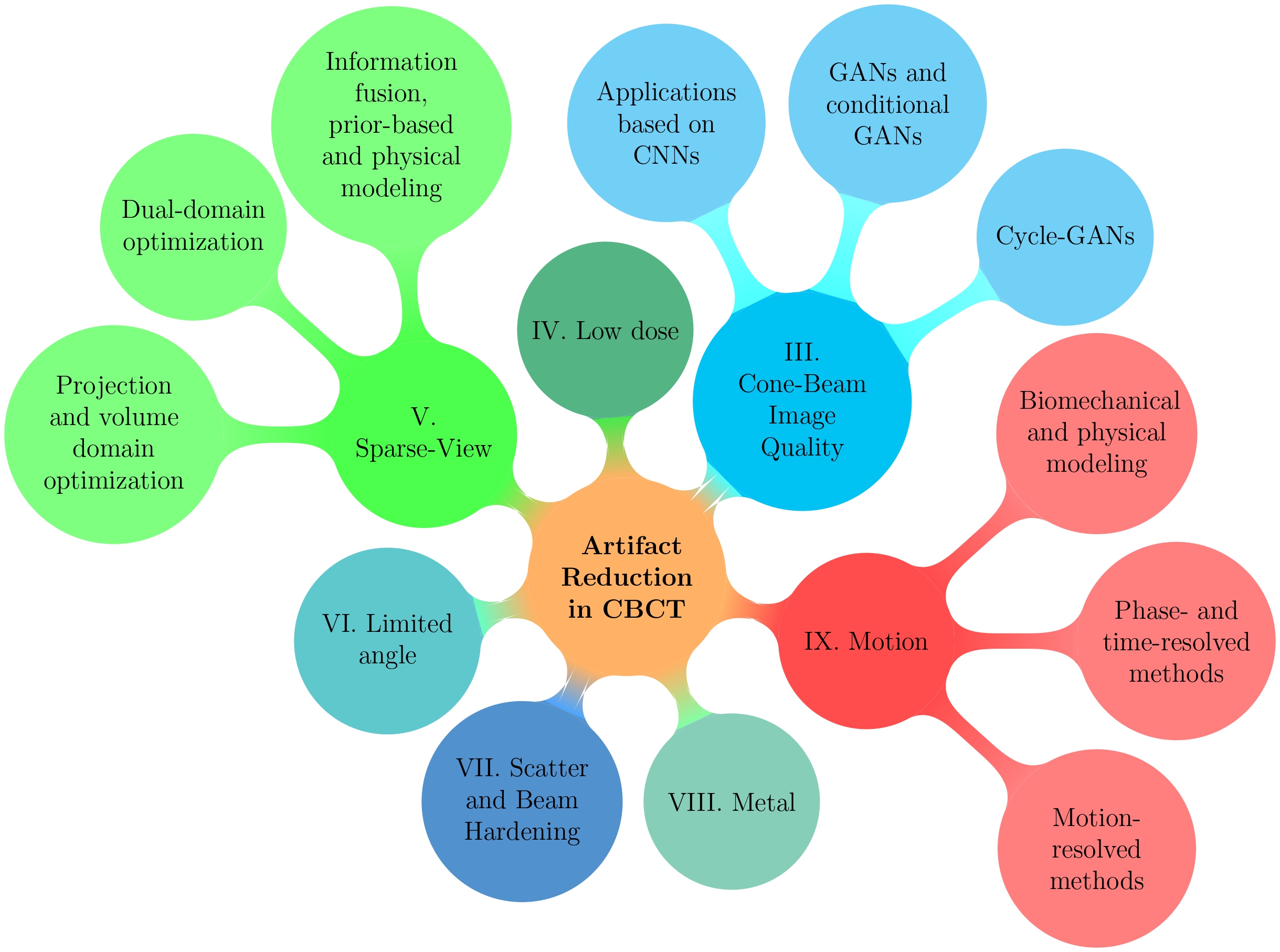}
    \caption{\hl{Visualisation of the content of this survey and the literature covered.}}
    \label{figure:content}
\end{figure}

\hl{This survey organizes the literature according to the type of artifacts which is addressed, and presents and contrasts the methodologies used within each specific artifact group (see Figure }\ref{figure:content}\hl{).}
The remainder of this paper is organized as follows:
Section~\ref{sec:preliminaries} briefly summarizes the basic aspects of CBCT acquisition and the assessment of scan quality.
Thereafter, the literature is discussed based on different types of artifacts (as outlined in~\cite{Schulze_2011, boas2012}) as follows:
Section~\ref{sec:image quality} presents methods attempting to improve CBCT image quality by reducing artifacts generated because of the cone-beam geometry and by bringing the CBCT quality closer to the one of CT scans.
The subsequent sections focus on various methods to address artifacts resulting from reduced acquisition dose.
Firstly, Section~\ref{sec:Low dose} discusses techniques that lower the dose per X-ray projection to achieve dose reduction. This is followed by Section~\ref{sec:sparse-view}, which explains methods for artifact reduction when acquiring fewer projections by uniformly dropping some of them (sparse-view reconstruction). 
Section~\ref{sec:limited angle} explores artifact reduction methods specifically for CBCT scans acquired from a limited angular range.
The paper then proceeds to discuss methods targeting scatter and beam hardening artifacts in Section~\ref{sec:scatter}. Section~\ref{sec:metal} is dedicated to research on reducing metal artifacts.
Section~\ref{sec:motion} focuses on motion compensation techniques for 3D and 4D CBCT. 
\hl{Further, the main trends in the recent literature on using deep learning-based architectures for CBCT artifact mitigation are presented in Section}~\ref{sec:discussions}\hl{, complemented with a discussion concerning the connections amongst the methods used for various types of artifacts and recommendations for future work.}
Finally, the paper concludes with Section~\ref{sec:conclusions}.

\section{Preliminaries}
\label{sec:preliminaries}

This section briefly reviews the basics of CBCT reconstruction and evaluation methods employed in artifact reduction and scan quality assessment.

\subsection{Cone-Beam Geometry Reconstruction and Deep Learning}

CBCT scans are acquired by means of an imaging system consisting of an X-ray source and a flat-panel (2D) detector mounted on a gantry system which rotates around the body region of interest. Several hundred 2D X-ray images are acquired at various angles.
These projections can be acquired from a limited angular range (so-called short scan) or a full $360^\circ$ trajectory (full scan).
Following the acquisition, a volumetric 3D image is reconstructed from the 2D projection images. Several methods exist to solve this ill-posed inverse problem. The most popular one is based on an analytic method developed by Feldkamp, Davis, and Kress (FDK~\cite{Feldkamp:84})  which provides a fast and reliable approximation of the inverse Radon transform. Alternatively, iterative algebraic reconstruction techniques (ART~\cite{gordon1970algebraic}) have become popular as well.
Moreover, by tracking the patients' motion, e.g. by capturing an external or internal breathing signal, and dividing the projections based on the motion state, it is possible to reconstruct 4D (motion-resolved) volumetric images. 4D scans include both the 3D volumetric information as well as their temporal dynamics.

In a nutshell, deep learning based approaches can be deployed at various stages of the CBCT reconstruction process. Firstly, deep neural networks can be trained to correct the acquired 2D projections (projection domain correction); secondly, they can be used to correct the reconstructed CBCT volumetric images (volume domain correction); and thirdly, the two approaches can be combined into a dual-domain correction.  Another approach is to augment or replace (parts of) the 2D-3D CBCT reconstruction itself with deep learning based components.
The components of the FDK algorithm were mapped into a deep neural network by means of a novel deep learning enabled cone beam back-projection layer~\cite{wurfl2018,maier2019nature}. The backward pass of the layer is computed as a forward projection operation. This approach thus permits joint optimization of correction steps in both volume and projection domain.  An open source implementation of differentiable reconstruction functions is available~\cite{syben_technical_2019}.
The networks are often trained in a supervised fashion by comparing reconstructed CBCT images with an artifact-free ground truth. \hl{Unsupervised}~\cite{Dong2022, Liu2022} \hl{and self-supervised}~\cite{Choi2022, Han2022} learning approaches have been employed as well. 

While datasets of 3D or 4D CBCT scans obtained from phantoms, animals or human subjects are available for training, they generally lack ground truth information required for deep learning based artifact mitigation employing supervised learning. To overcome this, artificial or simulated CBCT data is often used, obtained e.g. by means of forward projecting existing CT scans in a CBCT setup and manual incorporation of artifacts. For example, motion artifacts can be included by sampling CBCT projections at scan angles and time steps matching interpolated phases of a given 4D CT scan.

The general acquisition and reconstruction process of CBCT scans, including deep learning based corrections, is summarized in the visual abstract in Fig\hl{ure}~\ref{fig:visual_abstract}.

\subsection{Evaluation Metrics}

Several metrics have been utilized in the literature to evaluate the quality of CBCT scans enhanced by deep learning-based techniques.
The main qualitative evaluation metrics, computed between a reconstructed volume (with artifacts) and the ground truth reference, can be divided into two main groups as follows, according to~\cite{manual_rusanov_review_2022}:
\begin{itemize}
	\item \textbf{Image Similarity Metrics}: These metrics compute the similarity between scans and include (mean) absolute error (ME and MAE), (root) mean squared error (MSE and RMSE), (peak) signal-to-noise ratio (SNR and PSNR), structural similarity (SSIM)~\cite{wang2004image}, and Dice coefficient~\cite{dice}.
	\item \textbf{Dosimetric Similarity Metrics}: These metrics measure the consistency in dosimetry using a pair of scans, such as dose difference pass rate (DPR); dose–volume histogram (DVH), and gamma pass rate (GPR).  
\end{itemize}
In addition to the metrics mentioned above, metal artifact index (MAI~\cite{zhu2019evaluation}), and streak index (SI~\cite{cao2018comparative}) have been used in the literature to measure the level of specific artifacts in CT and CBCT scans. For motion, visual information fidelity (VIF)~\cite{vif} or autofocus (sharpness) metrics have been employed, among others.

\subsection{Clinical Evaluation}

The numerical evaluation metrics mentioned above compute the similarity of the improved CBCT compared with a reference, or report the level of the presence of artifacts, scan sharpness, or other quality criteria.
Ideally, these metrics should reflect the scan quality; hence, they should correspond to the preference of the experts in using the scans in clinical routine.
However, it is essential to note likely inconsistencies between simulated (where ground truth references exist) and real-world clinical data, so clinical evaluations are necessary to ensure the applicability of the presented methods for practical applications.
A clinical evaluation can be conducted by completing surveys with experts such as medical doctors or radiation physicists to directly assess the level of artifacts and the performance of the artifact reduction techniques, and the applicability of the improved images in various clinical tasks such as dose calculation, soft-tissue segmentation, and patient positioning~\cite{amirian2023mitigation}.

\section{Cone-Beam Image Quality}
\label{sec:image quality}
Cone-beam geometry and the size of the flat-panel detector result in the coverage of larger body areas but at lower resolution and degradation in scan quality compared to fan-beam CT scan acquisition.
Consequently, significant attention and extensive research has been directed at improving the quality of CBCT scans, often referred to as removing cone-beam or geometry artifacts in the literature.
One of the initial approaches to enhance CBCT quality involves employing supervised learning and training a 39-layer deep convolutional neural network (CNN) to map input CBCT scans to the corresponding planning CT as ground truth (reference) volumes~\cite{Kida2018}. This mapping of CBCT images to match correpsonding CT images is often called synthetic CT (sCT) from CBCT.

\paragraph*{\textbf{Applications based on CNNs}} Researchers have explored several CNN-based architectures with various supervised training objectives to enhance CBCT quality. 
For instance, denoising has been targeted through solving the multi-agent consensus equilibrium (MACE) problem and multi-slice information fusion techniques~\cite{Majee2019}. 
CNN models have demonstrated the ability to reduce ring artifacts from flat-panel CBCT scans using pre-corrected and artifact-free scans as ground truth~\cite{Chang2019}. 
Geometric artifacts caused by misalignment of the CBCT system %for oral phantoms 
were reduced using a modified fully convolutional neural network (M-FCNN), without using any pooling layers~\cite{Xiao2019}.
A further approach used a 3D block-based residual encoder-decoder convolutional neural network (RED-CNN) architecture coupled with a bilateral 3D filter and a 2D-based Landweber iteration to successfully remove Poisson noise while preserving the image structure at tissue edges~\cite{Choi2019}.
Training 3D models using a multi-task learning objective improved the quality of CBCTs by producing high-quality synthetic CT (sCT) scans from noisy and artifact-ridden scans for segmenting organs-at-risk (OARs) ~\cite{Dahiya2021}.
Lately, using InceptionV3~\cite{szegedy2016rethinkinginceptionv3} as a backbone has proven beneficial in reducing the artifacts observed in CBCT short scans due to the misalignment of the detection plane around the z-axis~\cite{Fang2022}.

\paragraph*{\textbf{GANs and conditional GANs}} Researchers have used self-supervised and unsupervised techniques to eliminate the need for paired CBCT and CT scans in supervised learning and to consider anatomical changes between the acquisition of planning CT (pCT) and CBCT.
These techniques mainly involve training auto-encoders, (conditional) generative adversarial networks (GANs~\cite{goodfellow2020generative}), and cycle-consistent generative adversarial networks (Cycle-GANs~\cite{zhu2017unpaired}).
Combining auto-encoders and GANs as a complementary approach to reweighting in analytical and iterative reconstruction methods has improved the quality of CBCT scans~\cite{Clark2019}. 
Training conditional GANs has shown promising results in enhancing the quality of CBCT through style transfer, effectively removing artifacts and discrepancies between CBCT and pCT for average tumor localization~\cite{Wei2020} and adaptive therapy~\cite{Santhanam2021}. 
Moreover, a more advanced GAN variant called temporal coherent generative adversarial network (TecoGAN) also improves the quality of simulated 4D CBCT scans by considering the time dependencies and motion for quality enhancement~\cite{chu2020learning, Zhang2021b}.

\paragraph*{\textbf{Cycle-GANs}} Using Cycle-GANs for unpaired translation from CBCT to pCT has received significant attention among researchers.
Notably, Cycle-GANs have successfully generated high-quality synthetic CT scans from CBCT for various organs, including prostate~\cite{Kida2020}, lung~\cite{Usui2022}, and abdominal scans~\cite{Hase2021}.
\hl{A novel architecture inspired by contrastive unpaired translation} (CUT~\cite{park2020cut}), \hl{trained in an unsupervised manner, improves the quality of CBCT scans by addressing fringe artifacts and noise degradation for dose calculation in adaptive radiotherapy}~\cite{Dong2022}.
The combination of binary cross-entropy, gradient difference, and identity losses with Cycle-GANs has further improved the quality of head and neck CBCT scans~\cite{Joseph2022}.
Introducing the residual block concept in the implementation of Res-Cycle-GAN has demonstrated advancements in the quality of sCT scans~\cite{Harms2019}.
Moreover, researchers have explored the combination of a Cycle-GAN with classical image processing techniques~\cite{OHara2022} and U-Net~\cite{unet_2015} architectures~\cite{Liu2022} in two-step approaches.
These approaches aim to initially reduce artifacts and subsequently generate sCT scans to improve the quality.
Ultimately, researchers demonstrated that trained Cycle-GANs enhance the quality of CBCT scans and achieve high accuracy in volumetric-modulated arc photon therapy (VMAT)~\cite{Kurz2019}.

\paragraph*{\textbf{Alternative methods}} In addition to adopting mainstream trends and computer vision architectures for artifact reduction in CBCT scans, researchers have explored creative methods specifically tailored to CBCT reconstruction using deep learning and neural networks.
For instance, U-Nets have been optimized for spectral blending of independently reconstructed sagittal and coronal views to enhance the CBCT quality~\cite{Han2020}.
Neural networks have also been integrated into the core of the reconstruction algorithms in the Feldkamp, Davis and
Kress (FDK) technique to introduce the NN-FDK technique for CBCT quality improvement~\cite{Lagerwerf2020}.
Another novel architecture, known as the iterative reconstruction network (AirNet), incorporates several variants in selecting projections based on random-phase (RP), prior-guided (PG), and all-phases (AP) for reconstruction~\cite{Chen2020b}.
Geometry-guided deep learning (GDL~\cite{Lu2021}), and its multi-beamlet-based approach (GMDL~\cite{Lu2022}) are additional examples of leveraging deep learning to enhance the reconstruction geometry effectively.
Finally, CNNs have been employed to predict the quality of the scans and accordingly dynamically adapt the C-arm source trajectory in the imaging acquisition process to avoid generating artifacts in the final scans~\cite{Thies2020}.

\section{Low dose}
\label{sec:Low dose}
The reduction of the acquisition dose in CBCT scans, which leads to the \hl{increased} presence of artifacts, has been addressed through various approaches such as adjusting the radiation dose per X-ray projection~\cite{Kofler2020}, increasing the acquisition speed or collecting fewer projections~\cite{Kurosawa2020}.
Early research focused on low-dose artifact reduction primarily by removing artifacts in the volume domain using deep CNNs with U-Net architectures.
The studies demonstrated the potential of decreasing the overall radiation dose through both dose reduction methods mentioned above~\cite{Kofler2020, Kurosawa2020}.
Moreover, a combination of 2D and 3D concatenating convolutional encoder-decoder (CCE-3D) with a structural sensitive loss (SSL) was employed to denoise low-dose CBCT scans and remove artifacts in both projection and volume domains.
This approach showed promising results in improving the quality of CBCT scans based on several metrics, such as PSNR and SSIM, and with greater improvements reported in the projection domain compared with the volume domain~\cite{Jin2022}.
In addition, a CNN-based iterative reconstruction framework was integrated with a plug-and-play proximal gradient descent framework to leverage DL-based denoising algorithms and enhance CBCT reconstruction~\cite{Kurosawa2020}.
Training models inspired by self-supervised learning approaches for inpainting and denoising Poisson and Gaussian noise have shown promising results in removing low-dose artifacts~\cite{Choi2021}.
Similarly, models optimized for removing Gaussian noise and addressing view aliasing artifacts through 2D iterations with 3D kernels have been developed~\cite{Choi2021a}. 
Furthermore, researchers combined a non-subsampled contourlet transform (NSCT) and a Sobel filter with U-Net architectures, referred to as NCS-Unet, to improve the quality of low-dose CBCT scans by enhancing both low- and high-frequency components~\cite{Chen2022}.

\section{Sparse-View}
\label{sec:sparse-view}
\begin{figure*}
	\centering
	\resizebox{0.95\textwidth}{!}{
		\begin{tabular}{ccc}
			%\toprule
			\includegraphics[width=0.33\textwidth]{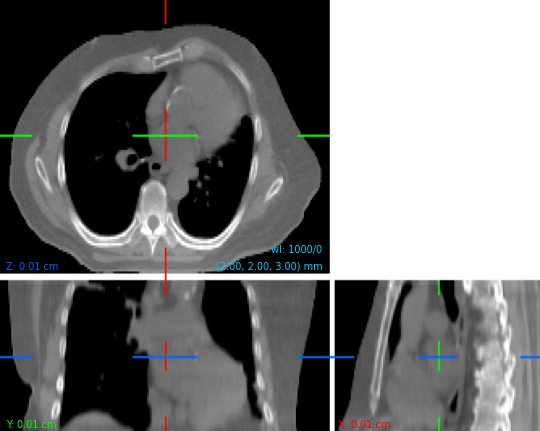} & 
			\includegraphics[width=0.33\textwidth]{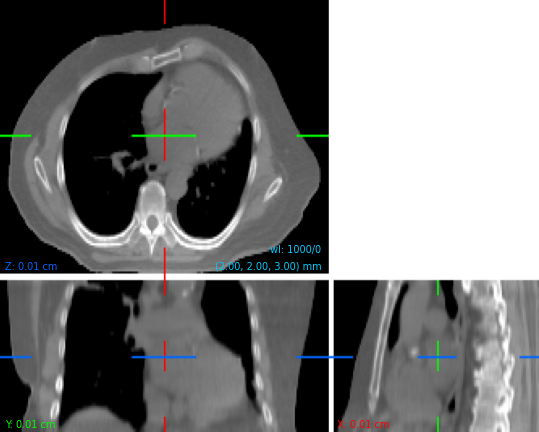} &
			\includegraphics[width=0.33\textwidth]{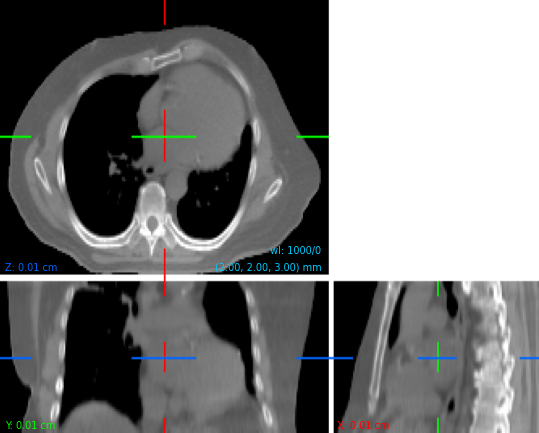} \\
			\multicolumn{3}{c}{Simulated 4D CBCT scan at three distinct motion phases, without significant motion artifacts} \\
			\midrule
			\includegraphics[width=0.33\textwidth]{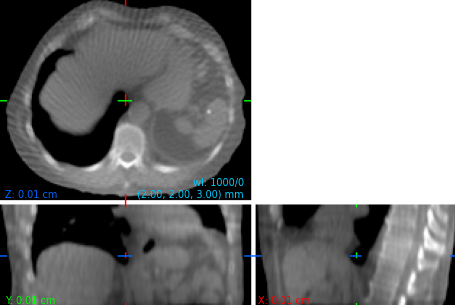} & 
			\includegraphics[width=0.33\textwidth]{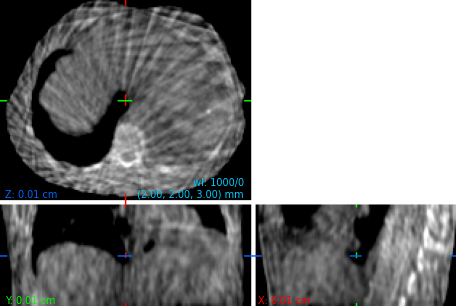} &
			\includegraphics[width=0.33\textwidth]{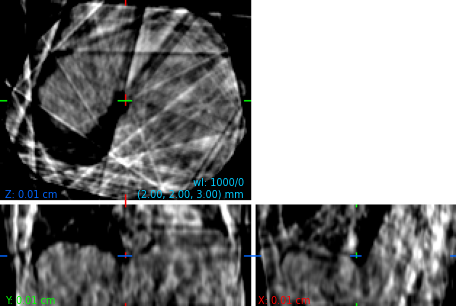} \\
			\multicolumn{3}{c}{Sparse-view artifacts at various sub-sampling rates (from left to right: 1/6, 1/18 and 1/48)}  \\
			\midrule
			& \\
			\includegraphics[height=0.20\textwidth]{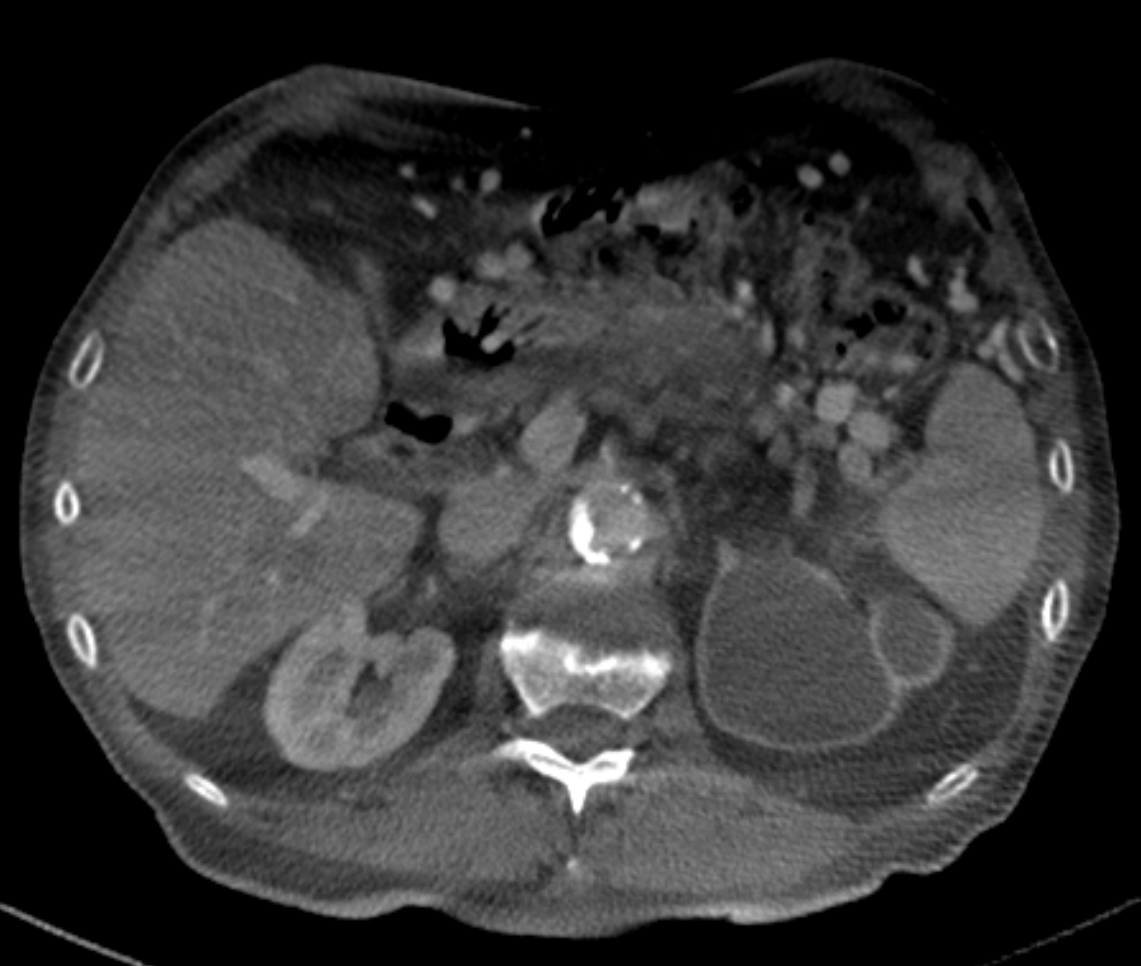} & 
			\includegraphics[height=0.20\textwidth]{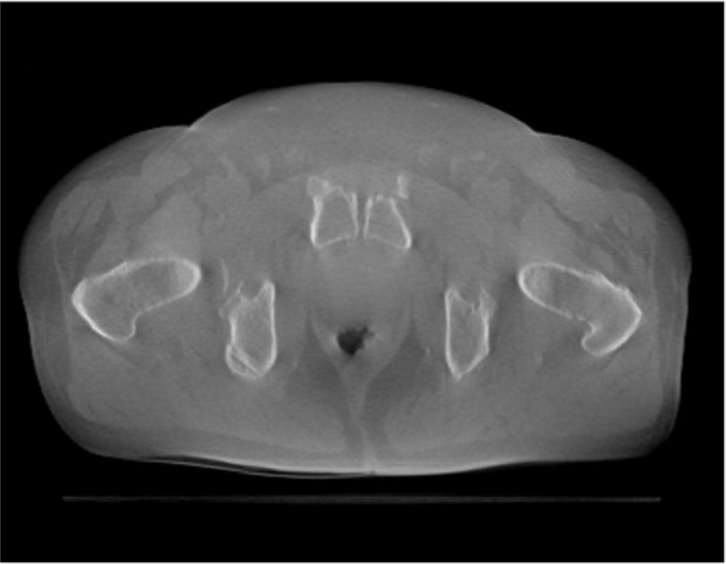} & 
			\includegraphics[height=0.20\textwidth]{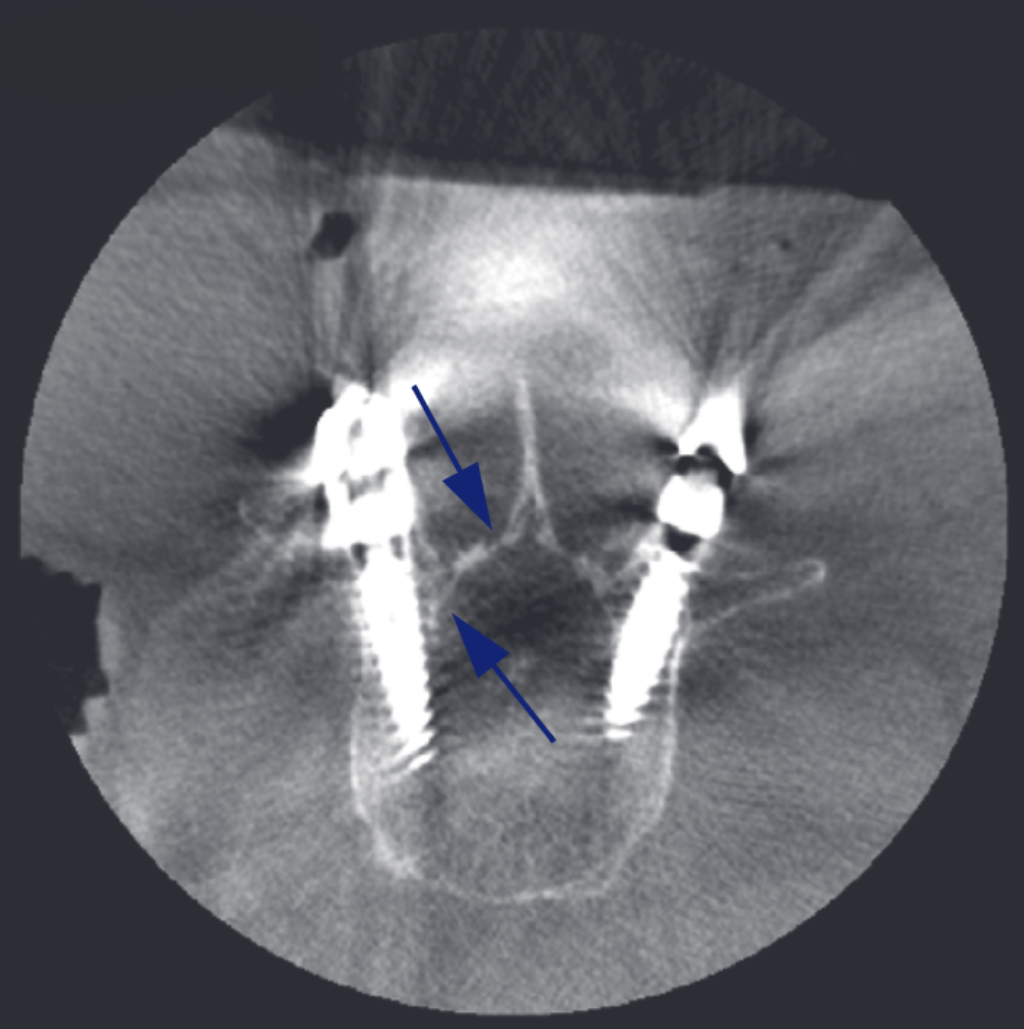} \\
			Limited angle artifacts~\cite{wurfl2018} & Scatter artifacts~\cite{Hansen2018} & Metal artifacts~\cite{Gottschalk2022} \\
			\midrule
			& \\
			\includegraphics[height=0.22\textwidth]{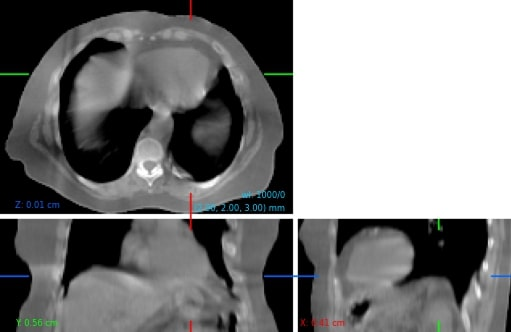} & 
			\includegraphics[height=0.22\textwidth]{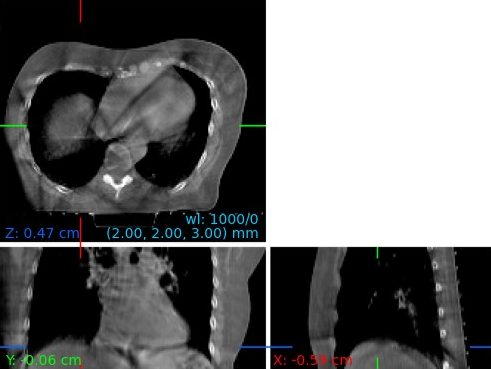} &
			\includegraphics[height=0.22\textwidth]{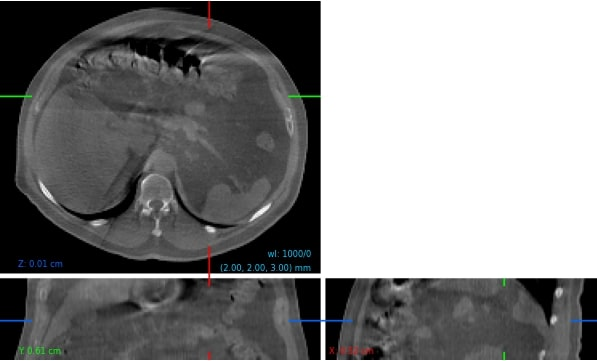} \\
			\multicolumn{3}{c}{Motion artifacts in simulated (left) and real (middle and right) CBCT scans~\cite{amirian2023mitigation}} \\
			%\toprule
		\end{tabular}
	}
	\caption{Examples of different kinds of artifacts appearing in CBCT scans. Shown are several artifact-free motion states obtained with a simulated 4D CBCT scan (1st row), sparse-view artifacts at various sub-sampling rates (2nd row), limited-angle, scatter and metal artifacts (3rd row), as well as motion artifacts (4th row).}
	\label{fig:sample_artifacts}
\end{figure*}

This section summarizes research aiming at reducing artifacts in CBCT reconstruction occurring from using uniformly downsampled full-scan ($360^{\circ}$) projections, primarily with the goal of dose reduction.
Sparse-view artifact reduction is closely related to \hl{mitigation of} artifacts caused by limited angle acquisition and breathing-phase-correlated 4D reconstruction, which will be reviewed in the upcoming sections~\ref{sec:limited angle} and~\ref{sec:motion}, respectively.
\hl{While the underlying motivations for sparse-view (acquisition dose reduction), limited angle (geometric constraints), and 4D (time resolved imaging) acquisition are different, in all cases artifacts are created due to the lack of projections from various angles.}
Decreasing the number of projections and the resulting data insufficiency for the reconstruction algorithm results in artifacts appearing in the shape of symmetric and uniform streaks, as depicted in Fig\hl{ure}~\ref{fig:sample_artifacts}.

\paragraph*{\textbf{Projection and volume domain optimization}} The body of literature on sparse-view artifact reduction using deep learning has been consistently growing since 2019, when initial research demonstrated the opportunity to reproduce the original image quality with using as few as one-seventh of the projections with symmetric CNN's as post-processing operation in the volume domain~\cite{Jiang2019}.
Similarly, using a multi-scale residual dense network (MS-RDN) successfully improved the quality of CBCTs reconstructed from one-third of the projections~\cite{Fu2020}.
In addition to training in the volume domain, the intensities of under-sampled projections can be corrected using deformation vector fields (DVFs) to match the original data, resulting in negligible streak artifacts after reconstruction~\cite{Zhang2019a}.
Similarly, symmetric residual CNN's (SR-CNN) can enhance the sharpness of the edges in anatomical structures reconstructed from sparse-view projections with total variation (TV) regularization in half-fan scans~\cite{Jiang2019}.
Furthermore, a counter-based total variational CBCT reconstruction using a U-Net architecture enhances the smoothed edges in lung CT reconstructed scans from half-fan projections~\cite{Chen2019}.
In Ref.~\cite{10.1007/978-3-031-26351-4_6}, a Reconstruction-Friendly Interpolation Network (RFI-Net) is developed, which uses a 3D-2D attention network to learn inter-projection relations for synthesizing missing projections, and then introduces a novel Ramp-Filter loss to constrain a frequency consistency between the synthesized and real projections.
The authors of \cite{Chao2022a} developed a dual-domain attention-guided network framework (Dual-AGNet) which works in both projection and reconstruction domains, featuring spatial attention modules and a joint loss function.

\paragraph*{\textbf{Dual-domain optimization}} Though interpolating missing data in the projections and removing artifacts in the volume domain are straightforward approaches to sparse-view artifact reduction, combining both and  backpropagating the error through the reconstruction algorithm is not trivial.
Despite the complexity involved, researchers attempted to unroll the proximal gradient descent algorithm for reconstruction and backpropagate the gradient through a U-Net architecture to reduce streak artifacts in~\cite{Wu2019}.
Since optimization in the volume domain and projection interpolation are regression problems with different or the same data channels as input and output, autoencoder-decoder architectures have also gained popularity for artifact reduction~\cite{Xie2019}.
To avoid complications regarding backpropagation through the reconstruction (back-projection) algorithm, DEER is introduced as an efficient end-to-end model for directly reconstructing CBCT scans from few-view projections~\cite{Xie2020}.
Furthermore, DeepOrganNet could fine-tune the lung mesh by skipping the reconstruction step and avoiding sparse-view artifacts appearing on organ mesh~\cite{Wang2020b}. \hl{Furthermore, the recent deep intensity field network (DIF-Net) model uses the latent representation (feature maps) of the 2D projections coupled with a view-specific query for extracting information from the projections. This information is then fed through cross-view fusion and intensity regression models to reconstruct a volume without artifacts.}~\cite{lin_learning_2023}.

\paragraph*{\textbf{Information fusion, prior-based and physical modeling}} Recent research trends seek to minimize sparse-view artifacts by incorporating multi-slice~\cite{Majee2021} and scale~\cite{Jiang2021} information fusion techniques, as well as combining information from different scan views (coronal, axial, and sagittal)~\cite{Junior2022}.
As the computational resources have become more powerful, deep learning for sparse-view artifact reduction has extended from 2D models for single slice processing to 3D models and processing of 4D CBCT scans~\cite{Majee2021}.
The use of prior (planing) CT and CBCT volumes to enhance the trained models, such as regularized iterative optimization reconstruction (PRIOR-Net~\cite{Hu2022}) and merge-encoder CNN (MeCNN~\cite{Jiang2021}) have recently become popular for sparse-view artifact reduction.
Researchers have also investigated using perception-aware~\cite{Ghosh2022} and physics-based~\cite{Hu2022} methods.
The learning paradigm has expanded beyond purely supervised learning 
to different tasks, such as denoising (DRUNet~\cite{devalla2018drunet}), artifact reduction~\cite{Jiang2022}, self-supervised by dropping projections~\cite{Han2022} and unsupervised learning through training conditional and generative adversarial networks (GANs)~\cite{Yang2022a}.

\section{Limited angle}
\label{sec:limited angle}

Besides lowering the imaging dose through uniformly downsampled projections, another approach to reducing the number of acquired projections and scanning dose is scanning the body from a limited angle.
Such scan settings are especially common when using a full-fan acquisition technique 
in a short-scan, where reconstruction is performed using projections from an angular range covering less than $360$ degrees. 
Although Parker weights~\cite{parkerweights} can be utilized to compensate for the loss of mass in the resulting CBCT scans, artifacts still appear due to the smaller number of acquired projections when scans are acquired from limited angles.
One of the initial attempts used learnable Parker weights
in the projection domain to address the mass loss in the angular range from $180^\circ + \theta$ to $360^\circ$ ($\theta$ being the fan angle)~\cite{wurfl2018}.
A subsequent study optimized a deep artifact correction model (DAC) using a 3D-ResUnet architecture to create high-quality scans and improve artifacts in limited-angle circular tomosynthesis (cTS), confirming the potential for quality enhancement in the volume domain~\cite{Schnurr2019}.
Further research demonstrated that combining FDK-based reconstruction with a neural network can achieve outstanding performance in 3D CBCT reconstruction from projections acquired from only $145^\circ$~\cite{Wang2020}.

Supervised learning, \hl{frequently} implemented through training U-Net architectures, for shading corrections in CBCT volumes with a narrow field of view (FOV) notably improved the quality of reconstructed CBCT scans, using CT scans as ground truth~\cite{Rossi2021}.
Another approach involves using a prior based on a fully sampled CT or CBCT and training a 2D3D-RegNet, which demonstrates the effectiveness of using a patient-specific prior for limited-angle sparseness artifact reduction~\cite{Zhang2021a}.
A conventional method for 4D CBCT reconstruction is dividing the projections based on the breathing phases and then reconstructing the body volume in those phases. 
As a result of using only a subset of the projections for each motion state, sparseness artifacts are prevalent for this special case of limited angle acquisition.
These artifacts have been addressed in the projection domain by interpolating the projections from different breathing phases~\cite{Dweek2022}.
In the volume domain, transfer learning, layer freezing, and fine-tuning have been employed to adapt the trained DL models to individual patients and mitigate sparseness artifacts~\cite{Sun2021}.

\section{Scatter and Beam Hardening}
\label{sec:scatter}
Large cone angles within the CBCT geometry setup have been observed to contribute to scatter artifacts, which have been addressed in the projection domain by leveraging Monte Carlo photon transport simulations to compute ground truth projections for supervised learning~\cite{Maier2018}.
A CNN-based deep scatter estimation (DSE~\cite{Maier2018}) architecture, as well as a  scatter correction network (ScatterNet~\cite{Hansen2018}) are the results of research endeavors using supervised learning for artifact correction in the projection domain.
The DSE model has demonstrated the potential to accurately emulate scatter artifacts and reduce the computational burden of using Monte-Carlo simulations while being orders of magnitude faster~\cite{Maier2019}. 
ScatterNet is considerably faster than the classical methods and might allow for on-the-fly shading correction~\cite{Hansen2018}.
ScatterNet, in combination with shading correction, also showed satisfactory results for dose calculation using volumetric modulated arc radiation therapy (VMAT), but yielded unsatisfactory outcomes for intensity-modulated proton therapy (IMPT).
Despite the abundant research work on scatter artifact corrections, studies tackling beam hardening are scarce. One such study involved training a U-Net-based architecture to predict monoenergetic X-ray projections from polyenergetic X-ray projections using  supervised learning on Monte Carlo simulation-based ground truth in the projection domain~\cite{Heyden2022}.

Compared with the classical fast adaptive scatter kernel superposition (fASKS) scatter reduction technique~\cite{Sun_2010}, a U-Net-based architecture outperformed in scatter artifact reduction for both full-fan and half-fan scans based on several metrics~\cite{Nomura2019}.
Additionally, a U-Net-based model trained on simulated CBCT projections has shown comparable performance to a validated empirical scatter correction technique in dose calculation for correcting the scatter artifacts in head and neck scans, computing the corrected volumes in less than 5 seconds~\cite{Lalonde2020}.
Besides classical approaches of scatter artifact reduction, CT scans have been used as ground truth volumes for training a modified U-Net architecture with a multi-objective loss function specifically targeting scatter artifact reduction in esophagus scans~\cite{Alam2021}. 

Apart from supervised learning methods, researchers have also trained Cycle-GAN models to improve the quality of CBCT scans, remove scatter artifacts, and generate sCT. In particular, Cycle-GAN has demonstrated superior performance compared to similar techniques using deep convolutional generative adversarial networks (DCGAN~\cite{liu2015faceattributes}) and  progressive growing GANs (PGGAN~\cite{karras2018progressive})~\cite{Liang2019}.

\section{Metal}
\label{sec:metal}

Metal objects and implants in the patient's body result in scattered radiation reaching the detector, leading to streak artifacts.
In the early research addressing metal artifacts, a CNN-based regression model has been trained to predict the detectability rank of metal implants to recommend out-of-plane angulation for C-arm source trajectories~\cite{Zaech2019}.
Further research in this area has proposed predicting the X-ray spectral shift after the localization of metal objects to define the optimal C-arm source-detector orbit~\cite{Wu2020a}.
The metal artifact avoidance (MAA) technique uses low-dose scout projections to roughly localize metal objects for the identification of a circular or non-circular orbit of C-arm source-detector to minimize variations in spectral shift and avoid metal artifacts~\cite{Wu2020}.

Researchers have also employed supervised learning for reducing metal artifacts and estimating the deviation of the voxel values after inserting neuroelectrodes~\cite{Uneri2021}.
Self-supervised learning approaches, focused on training models for inpainting the regions affected by metal artifacts, have demonstrated improvements in simultaneously tackling metal artifact reduction while preserving the essential anatomical structures near the inserted implants~\cite{Gottschalk2022}.
In addition to supervised and self-supervised techniques, various types of GANs have been employed in the literature for unsupervised metal artifact reduction. 
Optimized conventional GANs can reduce metal artifacts in high-resolution and physically realistic CT scans, with good generalization to clinical CBCT imaging technologies for inner-ear scans~\cite{Wang2022}. Conditional GANs, inspired by the pix2pix-GAN~\cite{isola2017image}, have successfully reduced metal artifacts in spine CBCT scans, enabling precise recovery of fiducial markers located outside the C-arm's field-of-view (FOV)~\cite{Fan2022}. A Cycle-GAN has also been employed to efficiently reduce metal artifacts by generating synthetic CT (sCT) from Megavolt CBCT (MVCBCT) and improving the quality of CBCT scans~\cite{Cao2022}.

\section{Motion}
\label{sec:motion}

Many of the state-of-the-art volumetric reconstruction techniques for CBCT rely heavily on the initial assumption that the projections are acquired from a stationary object. 
However, this assumption is often violated because of periodic respiratory and cardiac motions or non-voluntary and non-periodic movement of air bubbles in the abdominal area. 
When reconstructing CBCT volumes using projections acquired from various body states under motion, motion streak artifacts appear in the reconstructed volume, as shown in Fig\hl{ure}~\ref{fig:sample_artifacts}. The severity of the resulting artifacts is positively correlated with the intensity of motion.
The most common approach to tackle motion artifacts in CBCT scans is dividing the projections based on the motion state (motion-resolved~\cite{Beaudry2019, Lee2019a, Zhi2019, Madesta2020, Zhi2021, Zhi2021a}), periodic motion state (phase-resolved~\cite{Zhi2021, Zhang2022, Zhang2022c}) or acquisition time (time-resolved~\cite{Madesta2019, Thummerer2022}), and then reconstruct multiple volumes based on each batch of projections to generate a 4D CBCT. 

\paragraph*{\textbf{Motion-resolved methods}} A novel approach using CNNs to predict the missing projections in motion-resolved 4D-CBCT combined with a bin-sharing technique to accelerate the acquisition process, substantially removed streak artifacts compared with standard conjugate gradient reconstruction~\cite{Beaudry2019}.
Training a residual U-Net also reduces the streak artifacts appearing in 4D-CBCT by addressing the sparseness of the projections acquired in each breathing phase~\cite{Lee2019a}. 
Residual dense networks (RDNs~\cite{Madesta2020}) have successfully improved sparseness artifacts using an in-house lung and liver dataset, as well as a public dataset of the SPARE challenge~\cite{shieh2018aapm, shieh2019spare}.
Similar research demonstrates that combining the information of the different breathing phases to train a prior-guided CNN can effectively reduce artifacts in motion-resolved 4D-CBCT scans~\cite{Zhi2019}.
In addition to training single models, researchers attempted to optimize a cascade of spatial and temporal CNN models to combine spatial and temporal information for maximum artifact removal and to avoid errors in the tomographic information~\cite{Zhi2021a}.
A dual-encoder CNN (DeCNN) architecture simultaneously processes and combines the information of 4D motion-resolved volumes and the averaged volume, thereby improving the sharpness of the edges in moving and fixed tissues in 4D-CBCT~\cite{Jiang2022a}.

\paragraph*{\textbf{Phase- and time-resolved methods}} Phase-resolved CBCT is a specific case of motion-resolved CBCT, where projections are selected based on the different phases of body volume under periodic, respiratory, or cardiac motion. 
Motion Compensation Learning-induced sparse tensor constraint reconstruction (MCL-STCR) was shown to improve 4D-CBCT scans for all motion phases~\cite{Liu2019}.
3D-CNNs have shown to effectively mitigate sparse-view artifacts in motion-compensated 4D-CBCT scans reconstructed using FDK, thereby enhancing the overall quality~\cite{Zhang2022c}.
N-Net uses the prior volume reconstructed using all projections to remove streak artifacts. CycN-Net combines the temporal correlation among the phase-resolved scans to reduce streak artifacts that are caused by sparse-view sampled motion-resolved projections~\cite{Zhi2021}.
Furthermore, training a patient-specific GAN-based model on phase-resolved 4D-CBCT to reproduce CT quality using CBCT scans demonstrates improvements when applied to test set projections acquired from the same patient~\cite{Zhang2022}.
In addition to motion- and phase-resolved methods, training a U-Net can remove sparseness artifacts from time-resolved 4D-CBCT without requiring any prior information~\cite{Madesta2019}. 
GANs have also demonstrated the capacity of estimating sCT scans from time-resolved 4D-CBCT and the average 3D-CBCT volume, resulting in a comparable improvement in dose calculation using both strategies~\cite{Thummerer2022}.

\paragraph*{\textbf{Biomechanical and physical modeling}} In addition to phase-, motion-, and time-resolved techniques, researchers have also explored targeting motion artifacts by physically modeling the motion using a deformation-vector-field (DVF) and by optimizing an autofocus metric (i.e., maximizing some measure of sharpness).
The Simultaneous Motion Estimation and Image  Reconstruction (SMEIR) model, as well as its  biomechanical  modeling-guided version (SMEIR-Bio), are examples of models developed for motion effect prediction in lung 4D CBCT scans~\cite{Zhang2019}. 
These models have also been enhanced using a U-Net-based DVF optimization technique, leveraging a  population-based deep learning scheme to improve the accuracy of intra-lung DVF prediction  (SMEIR-Unet) in the same research work.
By incorporating the reference phase in 4D CBCT as an extra channel to their model, training a 4D U-Net for motion estimation, with fine-tuning the estimated DVFs, the performance of SMEIR models increases for motion artifact reduction~\cite{Huang2020}.
CNN-based architectures have been optimized to estimate deformable motion and predict the motion intensity on $8\times8$ grids covering the axial slice, followed by a preconditioning technique to favor more likely motion intensities~\cite{Sisniega2020}.
CNNs have also been trained for motion compensation in CBCT scans to solve the high-dimensional and no-convex problem of optimizing the autofocus metric~\cite{Sisniega2021}.

\paragraph*{\textbf{Alternative methods}} The autofocus metric has also been replaced with the Context-Aware Deep Learning-based Visual Information Fidelity (CADL-VIF) image similarity metric to optimize multi-resolution CNNs~\cite{huang2022}. 
This approach aims to improve motion degradation and compute sharp scans while preserving the tissue structures by optimizing visual information fidelity (VIF) without requiring motion-free ground truth.
An alternative to the autofocus metric is using contrastive loss to train GAN architectures to enhance the quality of 4D-CBCT scans and to reduce streak and motion artifacts~\cite{Dong2022}.
To address the slow speed of reconstruction and to compensate for the errors of 4D-CBCT due to the severe intraphase undersampling, a feature-compensated deformable convolutional network (FeaCo-DCN~\cite{Jiang2022b}) model has been proposed. It achieves nearly real-time reconstruction and accurate CBCT, outperforming the previous method applied to the SPARE Challenge~\cite{shieh2018aapm, shieh2019spare}. 
Besides the numerous research studies addressing motion in 4D CBCT, which requires recording the patient's breathing curve, researchers have also simulated motion in CBCT scans based on the estimation of DVFs according to 4D CT ground truth scans~\cite{herzig2022deep}. 
They subsequently trained a dual-domain model to mitigate 3D CBCT motion artifacts in the projection and volume domains. The clinical validation on real-world CBCT images yielded positive feedback from clinical experts, demonstrating the effectiveness of their approach for motion compensation~\cite{amirian2023mitigation}.
In addition to all methods to reduce motion artifacts, researchers have successfully used an artifact-driven slice sampling technique to avoid artifacts caused by moving air bubbles in the segmentation of the female pelvis~\cite{Hansch2019}.

\begin{figure}[htb!]
    \centering
    \subfloat[Distribution based on model architecture.]{
        \includegraphics[width=\linewidth]{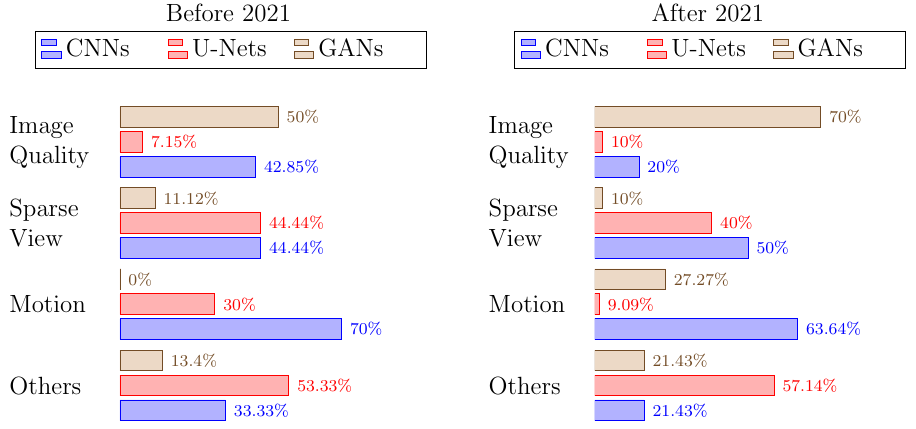}
        \label{fig:bar_models}
    }
    \hfill
    \subfloat[Distribution based on artifact type.]{
        \includegraphics[width=0.5\linewidth]{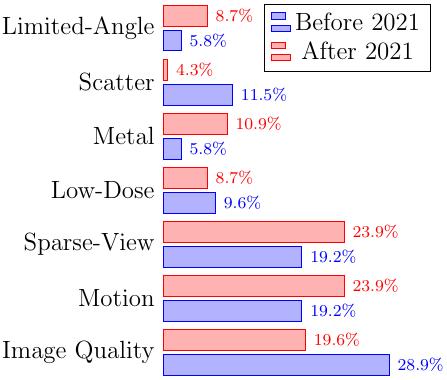}        
        \label{fig:bar_artifacts}
    }
    \caption{\hl{A visual summary of the distribution of the covered research literature in CBCT artifact mitigation using deep learning, separately for two time periods, (a) based on three generic deep learning architecture categories given a broad categorization by artifact type, and (b) based on the distribution according to the type of artifact.}}
%    \vspace{-0.6cm}
\end{figure}

\section{\hl{Discussion and Recommendations}}
\label{sec:discussions}

\begin{table*}[t]
    \centering
\footnotesize
\begin{tabular}{l|p{0.5cm}p{6.5cm}p{1.0cm}p{1.0cm}lp{1.6cm}p{1.0cm}}
    \hline
        Artifact type &  Year & Title & Anatomic site & Model & Patients & GPU Hardware & Published code? \\ \hline
        \multirow{2}{*}{image quality} & 2019 & Paired cycle‐GAN‐based image correction for quantitative cone‐beam computed tomography~\cite{Harms2019} & brain, pelvis & cycle GAN   & 44 & NVIDIA TITAN XP & - \\ \cline{2-8}
        & 2019 & CBCT correction using a cycle-consistent generative adversarial network and unpaired training to enable photon and proton dose calculation~\cite{Kurz2019} & pelvis & cycle GAN & 33 & NVIDIA Tesla P100 & - \\ \hline
        \multirow{2}{*}{low-dose} & 2019 & Computationally efficient deep neural network for computed tomography image reconstruction~\cite{Wu2019} & abdomen & U-Net & 10 & NVIDIA GTX 1080 Ti & - \\ \cline{2-8}
        & 2020 & Neural networks-based regularization for large-scale medical image reconstruction~\cite{Kofler2020} & cardiac & U-Net & 19 & - & - \\ \hline
        \multirow{4}{*}{sparse-view} & 2023 & Sub-volume-based Denoising Diffusion Probabilistic Model for Cone-beam CT Reconstruction from Incomplete Data~\cite{xia2023sub}& breast & diffusion model & - & 128x NVIDIA Tesla V100  & - \\ \cline{2-8}
        & 2023 & Learning Deep Intensity Field for Extremely Sparse-View CBCT Reconstruction ~\cite{lin_learning_2023}& knee & learned reconstruction & - & NVIDIA RTX 3090 & yes \\ \cline{2-8}
        & 2020 & Self‐contained deep learning‐based boosting of 4D cone‐beam CT reconstruction ~\cite{Madesta2020} & liver, lung & residual dense network & 20 & NVIDIA GeForce RTX 2080 Ti & yes \\ \cline{2-8}
        & 2020 & Deep Efficient End-to-End Reconstruction (DEER) Network for Few-View Breast CT Image Reconstruction ~\cite{Xie2020} & breast & GAN & 42 & NVIDIA Titan RTX & yes \\ \hline
        limited-angle & 2020 & C-arm orbits for metal artifact avoidance (MAA) in cone-beam CT~\cite{Wu2020} & chest phantom & U-Net & 0 & NVIDIA TITAN X & - \\ \hline
        scatter & 2019 & Real-time scatter estimation for medical CT using the deep scatter estimation: Method and robustness analysis with respect to different anatomies, dose levels, tube voltages, and data truncation~\cite{Maier2019} & head, thorax, pelvis & U-Net & 21 & NVIDIA Quadro P6000 & - \\ \hline
        metal & 2021 & Inner-ear augmented metal artifact reduction with simulation-based 3D generative adversarial networks~\cite{Wang2021} & temporal bone images & GAN & 597 & 11 GB GPU & - \\ \hline
        \multirow{2}{*}{motion} & 2022 & Enhancement of 4-D Cone-Beam Computed Tomography (4D-CBCT) Using a Dual-Encoder Convolutional Neural Network (DeCNN)~\cite{Jiang2022a} & lung & CNNs & 26 & NVIDIA Titan RTX & - \\ \cline{2-8}
        & 2022 & Deep learning‐based motion compensation for four‐dimensional cone‐beam computed tomography (4D‐CBCT) reconstruction~\cite{Zhang2022c} & thorax & CNNs & 18 & NVIDIA Tesla V100S & yes \\ \hline
    \end{tabular}
    \caption{\hl{Summary of a subset of studies selected guided by recency and number of citations. The table provides details about artifact category, publication year, study title, anatomic site, model type, number of patients, GPU hardware, and whether the code was published.}}
    \label{tab:papers}
\end{table*}

\hl{The previous sections have outlined the methodology and the complete workflow employed for deep learning based mitigation of artifacts in CBCT scans, addressing each specific type of artifact separately. This section presents a summary, emphasizing the central role of various deep learning approaches. The objective is to offer a comprehensive review of the architectures employed for different artifact types, highlighting both the promising aspects and the limitations in the current literature. 
%By doing so, it aims to indicate potential directions for future breakthroughs in this field.
}

\hl{
In general, a trend is observed in shifting from conventional supervised learning with CNNs and U-Net-type architectures to exploring more modern learning paradigms such as GANs, and investigating self-supervised and unsupervised methods, leveraging e.g. Cycle-GANs, as depicted in Figure}~\ref{fig:bar_models}.
\hl{In particular, Cycle-GAN-based architectures offer the appealing feature of enabling model training without needing paired labeled data}~\cite{Imae2020}.
\hl{However, they come with high data requirements, rising attention toward methods and projects for data collection, synthetical data generation, dataset merging from diverse sources, and data homogenization.
This trend suggests the rise of research works attempting at the adaptation of generative models including GANs, Cycle-GANs, as well as scored-based models}~\cite{song2022solving,wu_wavelet-improved_2023},
\hl{in upcoming research endeavors.
A recent example}~\cite{xia2023sub}\hl{, which employs denoising diffusion probabilistic models}~\cite{ddpm,kawar2022denoising}\hl{ for sparse-view CBCT reconstruction, demonstrates a lot of potential for future research, however at the expense of tremendous compute resources (up to 128 GPUs, see also Table}~\ref{tab:papers}). 
%\hl{While most models are trained on a single GPU, diffusion models can use up to 128 GPUs for optimization indicating a high computational cost for the training of these models}~\cite{xia2023sub}. 
\hl{On the other hand, less computationally intense, U-Net-based, architectures have demonstrated their merit in successfully addressing artifacts across all categories, making them a highly recommended and robust baseline approach for artifact mitigation.}

\hl{In the context of this survey, the primary DL-based architectures used in the literature can be divided into four key categories: CNNs, U-Nets, GANs, and cycle-GANs. Here, we categorize architectures with multi-scale information fusion, i.e.  including connections from the network's input (encoding) layers to output (decoding) layers (such as}~\cite{Wu2019})\hl{ under the category U-Net, while those without such direct connections (such as autoencoders}~\cite{Dai2019}) \hl{are categorized as CNNs.}
\hl{DL-based models generally require medium to large datasets for training, validation and testing through clinical evaluation. 
While medium-sized datasets, including multiple patients, can serve as starting points for training CNNs and U-Nets}~\cite{Rossi2021}\hl{, GANs perform better using datasets containing at least dozens of patient scans}~\cite{Hase2021}.
\hl{This trend generalizes to 3D and 4D reconstruction, where larger input sizes and a higher number of scans become essential, in particular for 4D}~\cite{Huang2020}.  \hl{A review of the studies presented in Table}~\ref{tab:papers} \hl{reveals that the majority of research was conducted with fewer than 50 patients. This relatively small number of patients can pose challenges for validating the approach across a diverse population. Consequently, the robustness of these models warrants further scrutiny to ensure their ability to generalize well across various human anatomies.} 

\hl{CNN architectures, known for their stable convergence and versatility, demonstrate a wide range of applications for artifact reduction through adapting different vision backbones}~\cite{Fang2022}\hl{ and incorporating diverse architectural components such as attention blocks}\cite{amirian2023mitigation}. 
\hl{However, in terms of multi-scale information fusion, they are inferior to U-Nets and their variants (e.g., U-Net++}~\cite{zhou_unet_2018})\hl{, which demonstrate a fast convergence in supervised learning due to the internal architectural connections between different layers enhancing the multi-resolution information fusion}~\cite{manual_rusanov_review_2022}.
\hl{Since CNNs and U-Nets are predominantly being trained in a supervised manner, their learning technique necessitates explicitly labeled data to define the task.
On the other hand, generative models (GANs), incorporating an adversarial loss, also offer potential applications in generating high-quality synthetic scans to meet the data needs of the deep learning-based architectures}~\cite{Wei2020}.
\hl{Moverover, Cycle-GANs compute the inverse path of artifact reduction automatically, using a cycle-consistent loss, thus being able to learn artifact reduction without the need for paired artifact-free ground truth}~\cite{Kurz2019}. 

\hl{Only four of the papers presented in Table} ~\ref{tab:papers} \hl{provide a public code repository to reproduce their results. This highlights a considerable shortage of open science practices, such as sharing code, to promote transparency and reproducibility in research. It is strongly recommended for researchers to share their code publicly to enhance the credibility and reproducibility of their work and accelerate scientific progress in this field.}

\section{Conclusions}
\label{sec:conclusions}

We presented a survey on the application of deep learning and convolutional neural networks to reduce various types of artifacts in CBCT scans. 
We categorized the existing literature based on the type of artifacts they address as well as the methodology employed.
Fig\hl{ure}~\ref{fig:bar_artifacts} illustrates the amount of the recent research works based on the type of artifacts.
It is observed that there has been considerable growth in artifact reduction research compared with focusing more generically on scan quality after 2021.
The opportunity of reducing the imaging dose with the help of compensating for artifacts when using low-dose scans, sparse-view, and limited-angle acquisition techniques have gained substantial attention due to the ease of simulation and computing the ground truth, especially for sparse-view and limited-angle approaches.

However, metal and scatter artifacts have received less attention. This may also be due to the challenges involved in computing the ground truth for metal artifacts, or the high computational cost of Monte-Carlo simulation for scatter artifacts.
We expect that the research community could profit from open-source accurate and fast artifact simulations for training models (as before with XCAT~\cite{segars20104d}).
The development of such simulations could also serve as a driving force for physics-based artifact modeling or training physics-informed neural networks (PINN)~\cite{pinn} for artifact reduction.
These simulations would benefit from GPU implementations for data generation to enable on-the-fly integration into the training pipelines with neural networks.
In addition to simulations, there is a research gap for open-source data augmentation techniques, such as~\cite{perez-garcia_torchio_2021, cardoso2022monai}, also based on incorporating simulated artifacts into real datasets.

In addition to simulation and augmentation tools for modelling, the research community would benefit from the availability of open-source datasets.
Researchers are still reporting results on phantoms and cadavers, indicating a need for more diverse and realistic publicly available datasets.
Nevertheless, despite the lack of open-source 4D CBCT datasets with raw projections and breathing curves, there is an increase of motion artifact reduction research in recent literature.
The collection and sharing of up-to-date benchmark datasets on a large scale, similar to the SPARSE~\cite{shieh2018aapm, shieh2019spare} and SynthRAD~\cite{adrian_thummerer_2023_7781049} challenges, would enhance the quality of many research works and provide the opportunity for fair and accurate comparison of different approaches.
Furthermore, many studies suffer from a lack of clinical evaluation. The availability of open-source standard clinical evaluation platforms would be of significant help in addressing this issue.

In terms of methodology, there has been a noticeable trend of moving beyond supervised learning towards self-supervised, unsupervised, and domain adaptation methods in recent years.
Researchers have started incorporating more physically inspired ideas into the neural networks and utilizing prior patient knowledge to personalize the models for specific anatomies.
One of the drawbacks often observed in the current literature is the absence of ablation studies. For example, in the case of approaches employing dual-domain optimization in both projection and volume domains,
the performance gained in each domain should be estimated separately. 
Besides artifact reduction after the CBCT acquisition, adapting the acquisition process itself using neural networks, such as C-arm trajectory adjustments applied to metal artifact reduction, present a further exciting avenue for future research.

In summary, substantial progress has been made in recent years transferring state-of-the-art methods from deep learning based computer vision to the domain of CBCT imaging and in particular the amelioration of prevalent imaging artifacts, with a clear potential to improve diagnosis and treatment in clinical practice.

%%%%%%%%%%%%%%%%%%%%%%%%%%%%%%%%%%%%%%%%%%%%%%%%%%%%%%%%%%%%%%%%%%%%%%%%%%%%%%%%%%%%%%%%%%%%%%%%%%%%%%%%%%

%\bibliographystyle{./IEEEtranDOI}
\bibliographystyle{IEEEtran}
\bibliography{IEEEabrv, ./references.bib, ./manual.bib}

\begin{thebibliography}{100}
\providecommand{\url}[1]{#1}
\csname url@samestyle\endcsname
\providecommand{\newblock}{\relax}
\providecommand{\bibinfo}[2]{#2}
\providecommand{\BIBentrySTDinterwordspacing}{\spaceskip=0pt\relax}
\providecommand{\BIBentryALTinterwordstretchfactor}{4}
\providecommand{\BIBentryALTinterwordspacing}{\spaceskip=\fontdimen2\font plus
\BIBentryALTinterwordstretchfactor\fontdimen3\font minus
  \fontdimen4\font\relax}
\providecommand{\BIBforeignlanguage}[2]{{%
\expandafter\ifx\csname l@#1\endcsname\relax
\typeout{** WARNING: IEEEtran.bst: No hyphenation pattern has been}%
\typeout{** loaded for the language `#1'. Using the pattern for}%
\typeout{** the default language instead.}%
\else
\language=\csname l@#1\endcsname
\fi
#2}}
\providecommand{\BIBdecl}{\relax}
\BIBdecl

\bibitem{Shende2016CommissioningOT}
R.~Shende, G.~Gupta, G.~Patel, and S.~Kumar, ``{Commissioning of TrueBeam(TM)
  Medical Linear Accelerator: Quantitative and Qualitative Dosimetric Analysis
  and Comparison of Flattening Filter (FF) and Flattening Filter Free (FFF)
  Beam},'' \emph{International Journal of Medical Physics, Clinical Engineering
  and Radiation Oncology}, vol.~5, pp. 51--69, 2016.

\bibitem{jaffray2002flat}
D.~A. Jaffray, J.~H. Siewerdsen, J.~W. Wong, and A.~A. Martinez, ``Flat-panel
  cone-beam computed tomography for image-guided radiation therapy,''
  \emph{International Journal of Radiation Oncology*Biology*Physics}, vol.~53,
  no.~5, pp. 1337--1349, 2002.

\bibitem{Goodfellow-et-al-2016}
\BIBentryALTinterwordspacing
I.~Goodfellow, Y.~Bengio, and A.~Courville, \emph{Deep Learning}.\hskip 1em
  plus 0.5em minus 0.4em\relax MIT Press, 2016. [Online]. Available:
  \url{http://www.deeplearningbook.org}
\BIBentrySTDinterwordspacing

\bibitem{paysan_annpr2020}
\BIBentryALTinterwordspacing
P.~Paysan, I.~Peterl{\'{\i}}k, T.~Roggen, L.~Zhu, C.~Wessels, J.~Schreier,
  M.~Buchacek, and S.~Scheib, ``{Deep Learning Methods for Image Guidance in
  Radiation Therapy},'' in \emph{Artificial Neural Networks in Pattern
  Recognition - 9th {IAPR} {TC3} Workshop, {ANNPR} 2020, Winterthur,
  Switzerland, September 2-4, 2020, Proceedings}, ser. Lecture Notes in
  Computer Science, F.-P. Schilling and T.~Stadelmann, Eds., vol. 12294.\hskip
  1em plus 0.5em minus 0.4em\relax Springer, 2020, pp. 3--22. [Online].
  Available: \url{https://doi.org/10.1007/978-3-030-58309-5\_1}
\BIBentrySTDinterwordspacing

\bibitem{spadea2021deep}
M.~F. Spadea, M.~Maspero, P.~Zaffino, and J.~Seco, ``{Deep learning based
  synthetic-CT generation in radiotherapy and PET: a review},'' \emph{Medical
  physics}, vol.~48, no.~11, pp. 6537--6566, 2021.

\bibitem{manual_zhang_review_2022}
\BIBentryALTinterwordspacing
M.~Zhang, S.~Gu, and Y.~Shi, ``The use of deep learning methods in low-dose
  computed tomography image reconstruction: a systematic review,''
  \emph{Complex \& Intelligent Systems}, vol.~8, no.~6, pp. 5545--5561, 2022.
  [Online]. Available: \url{https://doi.org/10.1007/s40747-022-00724-7}
\BIBentrySTDinterwordspacing

\bibitem{manual_rusanov_review_2022}
B.~Rusanov, G.~M. Hassan, M.~Reynolds, M.~Sabet, J.~Kendrick, P.~Rowshanfarzad,
  and M.~Ebert, ``{Deep learning methods for enhancing cone-beam CT image
  quality toward adaptive radiation therapy: A systematic review},''
  \emph{Medical Physics}, vol.~49, no.~9, pp. 6019--6054, 2022.

\bibitem{Schulze_2011}
R.~Schulze, U.~Heil, D.~Gro\ss, D.~Bruellmann, E.~Dranischnikow, U.~Schwanecke,
  and E.~Schoemer, ``Artefacts in {CBCT}: a review,'' \emph{Dentomaxillofacial
  Radiology}, vol.~40, no.~5, pp. 265--330, 2011.

\bibitem{boas2012}
F.~Boas and D.~Fleischmann, ``{CT artifacts: Causes and reduction
  techniques},'' \emph{Imaging in Medicine}, vol.~4, 2012.

\bibitem{Feldkamp:84}
L.~A. Feldkamp, L.~C. Davis, and J.~W. Kress, ``Practical cone-beam
  algorithm,'' \emph{J. Opt. Soc. Am. A}, vol.~1, no.~6, pp. 612--619, 1984.

\bibitem{gordon1970algebraic}
R.~Gordon, R.~Bender, and G.~T. Herman, ``{Algebraic reconstruction techniques
  (ART) for three-dimensional electron microscopy and x-ray photography},''
  \emph{Journal of Theoretical Biology}, vol.~29, no.~3, pp. 471--481, 1970.

\bibitem{wurfl2018}
T.~W{\"u}rfl, M.~Hoffmann, V.~Christlein, K.~Breininger, Y.~Huang, M.~Unberath,
  and A.~K. Maier, ``Deep learning computed tomography: Learning
  projection-domain weights from image domain in limited angle problems,''
  \emph{{IEEE} Transactions on Medical Imaging}, vol.~37, no.~6, pp.
  1454--1463, 2018.

\bibitem{maier2019nature}
A.~Maier, C.~Syben, B.~Stimpel, T.~Würfl, M.~Hoffmann, F.~Schebesch, W.~Fu,
  L.~Mill, L.~Kling, and S.~Christiansen, ``Learning with known operators
  reduces maximum error bounds,'' \emph{Nature Machine Intelligence}, vol.~1,
  pp. 373--380, 08 2019.

\bibitem{syben_technical_2019}
\BIBentryALTinterwordspacing
C.~Syben, M.~Michen, B.~Stimpel, S.~Seitz, S.~Ploner, and A.~K. Maier,
  ``Technical {Note}: {PYRO}‐{NN}: {Python} reconstruction operators in
  neural networks,'' \emph{Medical Physics}, vol.~46, no.~11, pp. 5110--5115,
  Nov. 2019. [Online]. Available:
  \url{https://www.ncbi.nlm.nih.gov/pmc/articles/PMC6899669/}
\BIBentrySTDinterwordspacing

\bibitem{Dong2022}
G.~Dong, C.~Zhang, L.~Deng, Y.~Zhu, J.~Dai, L.~Song, R.~Meng, T.~Niu, X.~Liang,
  and Y.~Xie, ``\BIBforeignlanguage{eng}{A deep unsupervised learning framework
  for the {4D} {CBCT} artifact correction.}''
  \emph{\BIBforeignlanguage{eng}{Physics in Medicine and Biology}}, vol.~67,
  no.~5, p. 055012, 2022.

\bibitem{Liu2022}
Y.~Liu, X.~Chen, J.~Zhu, B.~Yang, R.~Wei, R.~Xiong, H.~Quan, Y.~Liu, J.~Dai,
  and K.~Men, ``\BIBforeignlanguage{eng}{A two-step method to improve image
  quality of {CBCT} with phantom-based supervised and patient-based
  unsupervised learning strategies},'' \emph{\BIBforeignlanguage{eng}{Physics
  in medicine and biology}}, vol.~67, no.~8, 2022.

\bibitem{Choi2022}
{K. Choi}, ``A {Comparative} {Study} between {Image}- and {Projection}-{Domain}
  {Self}-{Supervised} {Learning} for {Ultra} {Low}-{Dose} {CBCT},'' in
  \emph{2022 44th {Annual} {International} {Conference} of the {IEEE}
  {Engineering} in {Medicine} \& {Biology} {Society} ({EMBC})}, 2022, pp.
  2076--2079.

\bibitem{Han2022}
Y.~Han and H.~Yu, ``Self-{Supervised} {Noise} {Reduction} in {Low}-{Dose}
  {Cone} {Beam} {Computed} {Tomography} ({CBCT}) {Using} the {Randomly}
  {Dropped} {Projection} {Strategy},'' \emph{Applied Sciences}, vol.~12, no.~3,
  p. 1714, 2022.

\bibitem{wang2004image}
Z.~Wang, A.~C. Bovik, H.~R. Sheikh, and E.~P. Simoncelli, ``Image quality
  assessment: from error visibility to structural similarity,'' \emph{IEEE
  Transactions on Image Processing}, vol.~13, no.~4, pp. 600--612, 2004.

\bibitem{dice}
L.~R. Dice, ``Measures of the amount of ecologic association between species,''
  \emph{Ecology}, vol.~26, no.~3, pp. 297--302, 1945.

\bibitem{zhu2019evaluation}
L.~Zhu, Y.~Chen, J.~Yang, X.~Tao, and Y.~Xi, ``Evaluation of the dental
  spectral cone beam ct for metal artefact reduction,''
  \emph{Dentomaxillofacial Radiology}, vol.~48, no.~2, p. 20180044, 2019.

\bibitem{cao2018comparative}
W.~Cao, T.~Sun, G.~Fardell, B.~Price, and W.~Dewulf, ``Comparative performance
  assessment of beam hardening correction algorithms applied on simulated data
  sets,'' \emph{Journal of Microscopy}, vol. 272, no.~3, pp. 229--241, 2018.

\bibitem{vif}
H.~Sheikh and A.~Bovik, ``Image information and visual quality,'' \emph{IEEE
  Transactions on Image Processing}, vol.~15, no.~2, pp. 430--444, 2006.

\bibitem{amirian2023mitigation}
M.~Amirian, J.~A. Montoya-Zegarra, I.~Herzig, P.~Eggenberger~Hotz,
  L.~Lichtensteiger, M.~Morf, A.~Z{\"u}st, P.~Paysan, I.~Peterlik, S.~Scheib,
  R.~M. Füchslin, T.~Stadelmann, and F.-P. Schilling, ``Mitigation of
  motion-induced artifacts in cone beam computed tomography using deep
  convolutional neural networks,'' \emph{Medical Physics}, vol.~50, pp.
  6228--6242, 2023.

\bibitem{Kida2018}
S.~Kida, T.~Nakamoto, M.~Nakano, K.~Nawa, A.~Haga, J.~Kotoku, H.~Yamashita, and
  K.~Nakagawa, ``\BIBforeignlanguage{eng}{Cone {Beam} {Computed} {Tomography}
  {Image} {Quality} {Improvement} {Using} a {Deep} {Convolutional} {Neural}
  {Network}.}'' \emph{\BIBforeignlanguage{eng}{Cureus}}, vol.~10, no.~4, p.
  e2548, 2018.

\bibitem{Majee2019}
{S. Majee}, {T. Balke}, {C. A. J. Kemp}, {G. T. Buzzard}, and {C. A. Bouman},
  ``{4D} {X}-{Ray} {CT} {Reconstruction} using {Multi}-{Slice} {Fusion},'' in
  \emph{2019 {IEEE} {International} {Conference} on {Computational}
  {Photography} ({ICCP})}, 2019, pp. 1--8.

\bibitem{Chang2019}
S.~Chang, X.~Chen, J.~Duan, and X.~Mou, ``A hybrid ring artifact reduction
  algorithm based on {CNN} in {CT} images,'' in \emph{15th International
  Meeting on Fully Three-Dimensional Image Reconstruction in Radiology and
  Nuclear Medicine}, S.~Matej and S.~Metzler, Eds., vol. 11072, 2019, p.
  1107226.

\bibitem{Xiao2019}
K.~Xiao, Y.~Han, Y.~Xu, L.~Li, X.~Xi, H.~Bu, and B.~Yan,
  ``\BIBforeignlanguage{eng}{X-ray cone-beam computed tomography geometric
  artefact reduction based on a data-driven strategy.}''
  \emph{\BIBforeignlanguage{eng}{Applied optics}}, vol.~58, no.~17, pp.
  4771--4780, Jun. 2019.

\bibitem{Choi2019}
D.~Choi, J.~Kim, S.~Chae, B.~Kim, J.~Baek, A.~Maier, R.~Fahrig, H.~Park, and
  J.~Choi, ``Multidimensional {Noise} {Reduction} in {C}-arm {Cone}-beam {CT}
  via {2D}-based {Landweber} {Iteration} and {3D}-based {Deep} {Neural}
  {Networks},'' in \emph{Medical Imaging 2019: Physics of Medical Imaging},
  T.~Schmidt, G.~Chen, and H.~Bosmans, Eds., vol. 10948, 2019, p. 1094837.

\bibitem{Dahiya2021}
N.~Dahiya, S.~R. Alam, P.~Zhang, S.-Y. Zhang, T.~Li, A.~Yezzi, and S.~Nadeem,
  ``\BIBforeignlanguage{eng}{Multitask {3D} {CBCT}-to-{CT} translation and
  organs-at-risk segmentation using physics-based data augmentation.}''
  \emph{\BIBforeignlanguage{eng}{Medical physics}}, vol.~48, no.~9, pp.
  5130--5141, 2021.

\bibitem{szegedy2016rethinkinginceptionv3}
C.~Szegedy, V.~Vanhoucke, S.~Ioffe, J.~Shlens, and Z.~Wojna, ``{Rethinking the
  Inception Architecture for Computer Vision},'' \emph{Proceedings of the IEEE
  conference on computer vision and pattern recognition ({CVPR})}, pp.
  2818--2826, 2016.

\bibitem{Fang2022}
Z.~Fang, B.~Ye, B.~Yuan, T.~Wang, S.~Zhong, S.~Li, and J.~Zheng, ``Angle
  prediction model when the imaging plane is tilted about z-axis,'' \emph{The
  Journal of Supercomputing}, vol.~78, no.~17, pp. 18\,598--18\,615, 2022.

\bibitem{goodfellow2020generative}
I.~Goodfellow, J.~Pouget-Abadie, M.~Mirza, B.~Xu, D.~Warde-Farley, S.~Ozair,
  A.~Courville, and Y.~Bengio, ``Generative adversarial networks,''
  \emph{Communications of the ACM}, vol.~63, no.~11, pp. 139--144, 2020.

\bibitem{zhu2017unpaired}
J.-Y. Zhu, T.~Park, P.~Isola, and A.~A. Efros, ``Unpaired image-to-image
  translation using cycle-consistent adversarial networks,'' in \emph{2017 IEEE
  International Conference on Computer Vision (ICCV)}, 2017, pp. 2242--2251.

\bibitem{Clark2019}
D.~Clark and C.~Badea, ``Spectral data completion for dual-source x-ray {CT},''
  in \emph{Medical Imaging 2019: Physics of Medical Imaging}, T.~Schmidt,
  G.~Chen, and H.~Bosmans, Eds., vol. 10948, 2019, p. 109481F.

\bibitem{Wei2020}
R.~Wei, B.~Liu, F.~Zhou, X.~Bai, D.~Fu, B.~Liang, and Q.~Wu,
  ``\BIBforeignlanguage{eng}{A patient-independent {CT} intensity matching
  method using conditional generative adversarial networks ({cGAN}) for single
  x-ray projection-based tumor localization.}''
  \emph{\BIBforeignlanguage{eng}{Physics in medicine and biology}}, vol.~65,
  no.~14, p. 145009, 2020.

\bibitem{Santhanam2021}
A.~Santhanam, M.~Lauria, B.~Stiehl, D.~Elliott, S.~Seshan, S.~Hsieh, M.~Cao,
  and D.~Low, ``An adversarial machine learning based approach and
  biomechanically-guided validation for improving deformable image registration
  accuracy between a planning {CT} and cone-beam {CT} for adaptive prostate
  radiotherapy applications,'' in \emph{Medical Imaging 2020: Image
  Processing}, I.~Isgum and B.~Landman, Eds., vol. 11313, 2021, p. 113130P.

\bibitem{chu2020learning}
M.~Chu, Y.~Xie, J.~Mayer, L.~Leal-Taix{\'e}, and N.~Thuerey, ``{Learning
  temporal coherence via self-supervision for GAN-based video generation},''
  \emph{ACM Transactions on Graphics}, vol.~39, no.~4, pp. 75:1--75:13, 2020.

\bibitem{Zhang2021b}
Z.~Zhang, M.~Huang, Z.~Jiang, Y.~Chang, J.~Torok, F.-F. Yin, and L.~Ren,
  ``\BIBforeignlanguage{eng}{{4D} radiomics: impact of {4D}-{CBCT} image
  quality on radiomic analysis.}'' \emph{\BIBforeignlanguage{eng}{Physics in
  Medicine and Biology}}, vol.~66, no.~4, p. 045023, 2021.

\bibitem{Kida2020}
S.~Kida, S.~Kaji, K.~Nawa, T.~Imae, T.~Nakamoto, S.~Ozaki, T.~Ohta, Y.~Nozawa,
  and K.~Nakagawa, ``\BIBforeignlanguage{eng}{Visual enhancement of {Cone}-beam
  {CT} by use of {CycleGAN}.}'' \emph{\BIBforeignlanguage{eng}{Medical
  physics}}, vol.~47, no.~3, pp. 998--1010, 2020.

\bibitem{Usui2022}
K.~Usui, K.~Ogawa, M.~Goto, Y.~Sakano, S.~Kyougoku, and H.~Daida,
  ``\BIBforeignlanguage{eng}{A cycle generative adversarial network for
  improving the quality of four-dimensional cone-beam computed tomography
  images},'' \emph{\BIBforeignlanguage{eng}{Radiation Oncology}}, vol.~17,
  p.~69, 2022.

\bibitem{Hase2021}
T.~Hase, M.~Nakao, K.~Imanishi, M.~Nakamura, T.~Matsuda, and {IEEE},
  ``Improvement of {Image} {Quality} of {Cone}-beam {CT} {Images} by
  {Three}-dimensional {Generative} {Adversarial} {Network},'' in \emph{2021
  43rd Annual International Conference of the IEEE Engineering in Medicine \&
  Biology Society (EMBC)}, 2021, pp. 2843--2846.

\bibitem{park2020cut}
T.~Park, A.~A. Efros, R.~Zhang, and J.-Y. Zhu, ``Contrastive learning for
  unpaired image-to-image translation,'' in \emph{European conference on
  computer vision ({ECCV})}, 2020, pp. 319--345.

\bibitem{Joseph2022}
J.~Joseph, P.~P.~N., and J.~P.~B., ``Supervised {Fan} {Beam} {Computed}
  {Tomography} {Image} {Synthesis} using {3D} {CycleGAN},'' in \emph{2022
  {IEEE} {International} {Conference} on {Signal} {Processing}, {Informatics},
  {Communication} and {Energy} {Systems} ({SPICES})}, vol.~1, 2022, pp. 81--86.

\bibitem{Harms2019}
J.~Harms, Y.~Lei, T.~Wang, R.~Zhang, J.~Zhou, X.~Tang, W.~J. Curran, T.~Liu,
  and X.~Yang, ``\BIBforeignlanguage{eng}{Paired cycle-{GAN}-based image
  correction for quantitative cone-beam computed tomography.}''
  \emph{\BIBforeignlanguage{eng}{Medical physics}}, vol.~46, no.~9, pp.
  3998--4009, 2019.

\bibitem{OHara2022}
C.~J. O'Hara, D.~Bird, B.~Al-Qaisieh, and R.~Speight,
  ``\BIBforeignlanguage{eng}{Assessment of {CBCT}-based synthetic {CT}
  generation accuracy for adaptive radiotherapy planning.}''
  \emph{\BIBforeignlanguage{eng}{Journal of applied clinical medical physics}},
  vol.~23, no.~11, p. e13737, 2022.

\bibitem{unet_2015}
O.~Ronneberger, P.~Fischer, and T.~Brox, ``U-net: Convolutional networks for
  biomedical image segmentation,'' in \emph{Medical Image Computing and
  Computer-Assisted Intervention -- MICCAI 2015}, N.~Navab, J.~Hornegger, W.~M.
  Wells, and A.~F. Frangi, Eds.\hskip 1em plus 0.5em minus 0.4em\relax Cham:
  Springer International Publishing, 2015, pp. 234--241.

\bibitem{Kurz2019}
C.~Kurz, M.~Maspero, M.~H.~F. Savenije, G.~Landry, F.~Kamp, M.~Pinto, M.~Li,
  K.~Parodi, C.~Belka, and C.~A.~T. van~den Berg,
  ``\BIBforeignlanguage{eng}{{CBCT} correction using a cycle-consistent
  generative adversarial network and unpaired training to enable photon and
  proton dose calculation},'' \emph{\BIBforeignlanguage{eng}{Physics in
  medicine and biology}}, vol.~64, no.~22, p. 225004, 2019.

\bibitem{Han2020}
{Y. Han}, {J. Kim}, and {J. C. Ye}, ``Differentiated {Backprojection} {Domain}
  {Deep} {Learning} for {Conebeam} {Artifact} {Removal},'' \emph{IEEE
  Transactions on Medical Imaging}, vol.~39, no.~11, pp. 3571--3582, 2020.

\bibitem{Lagerwerf2020}
M.~J. Lagerwerf, D.~M. Pelt, W.~J. Palenstijn, and K.~J. Batenburg,
  ``\BIBforeignlanguage{eng}{A {Computationally} {Efficient} {Reconstruction}
  {Algorithm} for {Circular} {Cone}-{Beam} {Computed} {Tomography} {Using}
  {Shallow} {Neural} {Networks}.}'' \emph{\BIBforeignlanguage{eng}{Journal of
  imaging}}, vol.~6, no.~12, p. 135, 2020.

\bibitem{Chen2020b}
G.~Chen, Y.~Zhao, Q.~Huang, and H.~Gao, ``{4D}-{AirNet}: a temporally-resolved
  {CBCT} slice reconstruction method synergizing analytical and iterative
  method with deep learning,'' \emph{Physics in Medicine and Biology}, vol.~65,
  no.~17, 2020.

\bibitem{Lu2021}
K.~Lu, L.~Ren, and F.-F. Yin, ``\BIBforeignlanguage{eng}{A geometry-guided deep
  learning technique for {CBCT} reconstruction.}''
  \emph{\BIBforeignlanguage{eng}{Physics in Medicine and Biology}}, vol.~66,
  no.~15, p. 15LT01, 2021.

\bibitem{Lu2022}
------, ``\BIBforeignlanguage{eng}{A geometry-guided multi-beamlet deep
  learning technique for {CT} reconstruction.}''
  \emph{\BIBforeignlanguage{eng}{Biomedical Physics \& Engineering Express}},
  vol.~8, no.~4, p. 045004, 2022.

\bibitem{Thies2020}
M.~Thies, J.-N. Zäch, C.~Gao, R.~Taylor, N.~Navab, A.~Maier, and M.~Unberath,
  ``\BIBforeignlanguage{eng}{A learning-based method for online adjustment of
  {C}-arm {Cone}-beam {CT} source trajectories for artifact avoidance},''
  \emph{\BIBforeignlanguage{eng}{International journal of computer assisted
  radiology and surgery}}, vol.~15, no.~11, pp. 1787--1796, 2020.

\bibitem{Kofler2020}
A.~Kofler, M.~Haltmeier, T.~Schaeffter, M.~Kachelrieß, M.~Dewey, C.~Wald, and
  C.~Kolbitsch, ``\BIBforeignlanguage{eng}{Neural networks-based regularization
  for large-scale medical image reconstruction.}''
  \emph{\BIBforeignlanguage{eng}{Physics in Medicine and Biology}}, vol.~65,
  no.~13, p. 135003, 2020.

\bibitem{Kurosawa2020}
T.~Kurosawa, T.~Nishio, S.~Moriya, M.~Tsuneda, and K.~Karasawa,
  ``\BIBforeignlanguage{eng}{Feasibility of image quality improvement for
  high-speed {CBCT} imaging using deep convolutional neural network for
  image-guided radiotherapy in prostate cancer.}''
  \emph{\BIBforeignlanguage{eng}{Physica Medica}}, vol.~80, pp. 84--91, 2020.

\bibitem{Jin2022}
D.~S.-C. Jin, L.-S. Chang, Y.-H. Wang, J.-C. Chen, S.~H. Tseng, and T.-Y. Liu,
  ``\BIBforeignlanguage{eng}{Virtual and real-world implementation of
  deep-learning-based image denoising model on projection domain in digital
  tomosynthesis and cone-beam computed tomography data},''
  \emph{\BIBforeignlanguage{eng}{Biomedical physics \& engineering express}},
  vol.~8, no.~6, 2022.

\bibitem{Choi2021}
{K. Choi}, ``Self-supervised {Projection} {Denoising} for {Low}-{Dose}
  {Cone}-{Beam} {CT},'' in \emph{2021 43rd {Annual} {International}
  {Conference} of the {IEEE} {Engineering} in {Medicine} \& {Biology} {Society}
  ({EMBC})}, Nov. 2021, pp. 3459--3462.

\bibitem{Choi2021a}
D.~Choi, W.~Kim, J.~Lee, M.~Han, J.~Baek, and J.~Choi, ``Integration of {2D}
  iteration and a {3D} {CNN}-based model for multi-type artifact suppression in
  {C}-arm cone-beam {CT},'' \emph{Machine Vision and Applications}, vol.~32,
  no. 116, 2021.

\bibitem{Chen2022}
K.~Chen, L.~Zhang, J.~Liu, Y.~Gao, Z.~Wu, H.~Zhu, C.~Du, X.~Mai, C.~Yang, and
  Y.~Chen, ``Robust restoration of low-dose cerebral perfusion {CT} images
  using {NCS}-{Unet},'' \emph{Nuclear Science and Techniques}, vol.~33, no.~30,
  2022.

\bibitem{Jiang2019}
{Z. Jiang}, {Y. Chen}, {Y. Zhang}, {Y. Ge}, {F. -F. Yin}, and {L. Ren},
  ``Augmentation of {CBCT} {Reconstructed} {From} {Under}-{Sampled}
  {Projections} {Using} {Deep} {Learning},'' \emph{IEEE Transactions on Medical
  Imaging}, vol.~38, no.~11, pp. 2705--2715, 2019.

\bibitem{Fu2020}
Z.~Fu, H.~Tseng, S.~Vedantham, A.~Karellas, and A.~Bilgin, ``A residual dense
  network assisted sparse view reconstruction for breast computed tomography,''
  \emph{scientific reports}, vol.~10, p. 21111, 2020.

\bibitem{Zhang2019a}
Y.~Zhang, L.~Chen, B.~Li, M.~Folkert, X.~Jia, X.~Gu, and J.~Wang,
  ``Incorporating {Biomechanical} {Modeling} and {Deep} {Learning} into a
  {Deformation}-{Driven} {Liver} {CBCT} {Reconstruction} {Technique},'' in
  \emph{Medical Imaging 2019: Physics of Medical Imaging}, T.~Schmidt, G.~Chen,
  and H.~Bosmans, Eds., vol. 10948, 2019.

\bibitem{Chen2019}
Y.~Chen, F.-F. Yin, Z.~Jiang, and L.~Ren, ``\BIBforeignlanguage{eng}{Daily edge
  deformation prediction using an unsupervised convolutional neural network
  model for low dose prior contour based total variation {CBCT} reconstruction
  ({PCTV}-{CNN})},'' \emph{\BIBforeignlanguage{eng}{Biomedical Physics \&
  Engineering Express}}, vol.~5, no.~6, p. 065013, 2019.

\bibitem{10.1007/978-3-031-26351-4_6}
Y.~Wang, L.~Chao, W.~Shan, H.~Zhang, Z.~Wang, and Q.~Li, ``Improving the
  quality of sparse-view cone-beam computed tomography via
  reconstruction-friendly interpolation network,'' in \emph{Computer Vision --
  ACCV 2022}, L.~Wang, J.~Gall, T.-J. Chin, I.~Sato, and R.~Chellappa,
  Eds.\hskip 1em plus 0.5em minus 0.4em\relax Cham: Springer Nature
  Switzerland, 2023, pp. 86--100.

\bibitem{Chao2022a}
L.~Chao, P.~Zhang, Y.~Wang, Z.~Wang, W.~Xu, and Q.~Li, ``Dual-domain
  attention-guided convolutional neural network for low-dose cone-beam computed
  tomography reconstruction,'' \emph{Knowledge-Based Systems}, vol. 251, Sep.
  2022.

\bibitem{Wu2019}
D.~Wu, K.~Kim, and Q.~Li, ``\BIBforeignlanguage{eng}{Computationally efficient
  deep neural network for computed tomography image reconstruction},''
  \emph{\BIBforeignlanguage{eng}{Medical Physics}}, vol.~46, no.~11, pp.
  4763--4776, 2019.

\bibitem{Xie2019}
H.~Xie, H.~Shan, and G.~Wang, ``Deep {Encoder}-{Decoder} {Adversarial}
  {Reconstruction} ({DEAR}) {Network} for {3D} {CT} from {Few}-{View} {Data},''
  \emph{Bioengineering}, vol.~6, no.~4, 2019.

\bibitem{Xie2020}
{H. Xie}, {H. Shan}, {W. Cong}, {C. Liu}, {X. Zhang}, {S. Liu}, {R. Ning}, and
  {G. Wang}, ``Deep {Efficient} {End}-to-{End} {Reconstruction} ({DEER})
  {Network} for {Few}-{View} {Breast} {CT} {Image} {Reconstruction},''
  \emph{IEEE Access}, vol.~8, pp. 196\,633--196\,646, 2020.

\bibitem{Wang2020b}
Y.~Wang, Z.~Zhong, and J.~Hua, ``\BIBforeignlanguage{eng}{{DeepOrganNet}:
  {On}-the-{Fly} {Reconstruction} and {Visualization} of {3D} / {4D} {Lung}
  {Models} from {Single}-{View} {Projections} by {Deep} {Deformation}
  {Network}},'' \emph{\BIBforeignlanguage{eng}{IEEE transactions on
  visualization and computer graphics}}, vol.~26, no.~1, pp. 960--970, 2020.

\bibitem{lin_learning_2023}
Y.~Lin, Z.~Luo, W.~Zhao, and X.~Li, ``\BIBforeignlanguage{en}{Learning {Deep}
  {Intensity} {Field} for {Extremely} {Sparse}-{View} {CBCT}
  {Reconstruction}},'' in \emph{\BIBforeignlanguage{en}{Medical {Image}
  {Computing} and {Computer} {Assisted} {Intervention} – {MICCAI} 2023}},
  ser. Lecture {Notes} in {Computer} {Science}, H.~Greenspan, A.~Madabhushi,
  P.~Mousavi, S.~Salcudean, J.~Duncan, T.~Syeda-Mahmood, and R.~Taylor,
  Eds.\hskip 1em plus 0.5em minus 0.4em\relax Cham: Springer Nature
  Switzerland, 2023, pp. 13--23.

\bibitem{Majee2021}
{S. Majee}, {T. Balke}, {C. A. J. Kemp}, {G. T. Buzzard}, and {C. A. Bouman},
  ``Multi-{Slice} {Fusion} for {Sparse}-{View} and {Limited}-{Angle} {4D} {CT}
  {Reconstruction},'' \emph{IEEE Transactions on Computational Imaging},
  vol.~7, pp. 448--462, 2021.

\bibitem{Jiang2021}
Z.~Jiang, Z.~Zhang, Y.~Chang, Y.~Ge, F.-F. Yin, and L.~Ren,
  ``\BIBforeignlanguage{eng}{Prior image-guided cone-beam computed tomography
  augmentation from under-sampled projections using a convolutional neural
  network.}'' \emph{\BIBforeignlanguage{eng}{Quantitative imaging in medicine
  and surgery}}, vol.~11, no.~12, 2021.

\bibitem{Junior2022}
C.~A.~A. Júnior, L.~F.~A. Pereira, G.~D.~C. Cavalcanti, and T.~I. Ren,
  ``Ensemble of {Convolutional} {Neural} {Networks} for {Sparse}-{View}
  {Cone}-{Beam} {Computed} {Tomography},'' in \emph{2022 {International}
  {Joint} {Conference} on {Neural} {Networks} ({IJCNN})}, 2022, pp. 1--7.

\bibitem{Hu2022}
{D. Hu}, {Y. Zhang}, {J. Liu}, {Y. Zhang}, {J. L. Coatrieux}, and {Y. Chen},
  ``{PRIOR}: {Prior}-{Regularized} {Iterative} {Optimization} {Reconstruction}
  {For} {4D} {CBCT},'' \emph{IEEE Journal of Biomedical and Health
  Informatics}, vol.~26, no.~11, pp. 5551--5562, 2022.

\bibitem{Ghosh2022}
{S. Ghosh}, {P. Ernst}, {G. Rose}, {A. Nürnberger}, and {S. Stober}, ``Towards
  {Patient} {Specific} {Reconstruction} {Using} {Perception}-{Aware} {CNN} and
  {Planning} {CT} as {Prior},'' in \emph{2022 {IEEE} 19th {International}
  {Symposium} on {Biomedical} {Imaging} ({ISBI})}, 2022, pp. 1--5.

\bibitem{devalla2018drunet}
S.~K. Devalla, P.~K. Renukanand, B.~K. Sreedhar, G.~Subramanian, L.~Zhang,
  S.~Perera, J.-M. Mari, K.~S. Chin, T.~A. Tun, N.~G. Strouthidis
  \emph{et~al.}, ``{DRUNET: a dilated-residual U-Net deep learning network to
  segment optic nerve head tissues in optical coherence tomography images},''
  \emph{Biomedical optics express}, vol.~9, no.~7, pp. 3244--3265, 2018.

\bibitem{Jiang2022}
{L. Jiang}, {X. Wang}, {H. Jiang}, {X. Wang}, {H. Guo}, and {X. He},
  ``Sparse-{View} {CBCT} {Reconstruction} {Using} {Combined} {DRUNet} and
  {HQS},'' in \emph{2022 5th {International} {Conference} on {Pattern}
  {Recognition} and {Artificial} {Intelligence} ({PRAI})}, 2022, pp.
  1051--1054.

\bibitem{Yang2022a}
Y.~Yang, C.~Fang, and L.~Zhu, ``\BIBforeignlanguage{chi}{{Sparse}-view
  {Cone}-beam {Breast} {CT} {Reconstruction} via {cGAN} {Constrained} by
  {Image} {Edges}},'' \emph{\BIBforeignlanguage{chi}{Zhongguo yi liao qi xie za
  zhi (Chinese journal of medical instrumentation)}}, vol.~46, no.~2, pp.
  119--125, 2022.

\bibitem{parkerweights}
D.~L. Parker, ``{Optimal short scan convolution reconstruction for fan beam
  CT},'' \emph{Medical Physics}, vol.~9, no.~2, pp. 254--257, 1982.

\bibitem{Schnurr2019}
A.-K. Schnurr, K.~Chung, T.~Russ, L.~R. Schad, and F.~G. Zöllner,
  ``\BIBforeignlanguage{eng}{Simulation-based deep artifact correction with
  {Convolutional} {Neural} {Networks} for limited angle artifacts.}''
  \emph{\BIBforeignlanguage{eng}{Zeitschrift f\"ur Medizinische Physik}},
  vol.~29, no.~2, pp. 150--161, 2019.

\bibitem{Wang2020}
{Y. Wang}, {T. Yang}, and {W. Huang}, ``Limited-{Angle} {Computed} {Tomography}
  {Reconstruction} using {Combined} {FDK}-{Based} {Neural} {Network} and
  {U}-{Net},'' in \emph{2020 42nd {Annual} {International} {Conference} of the
  {IEEE} {Engineering} in {Medicine} \& {Biology} {Society} ({EMBC})}, 2020,
  pp. 1572--1575.

\bibitem{Rossi2021}
M.~Rossi, G.~Belotti, C.~Paganelli, A.~Pella, A.~Barcellini, P.~Cerveri, and
  G.~Baroni, ``\BIBforeignlanguage{eng}{Image-based shading correction for
  narrow-{FOV} truncated pelvic {CBCT} with deep convolutional neural networks
  and transfer learning.}'' \emph{\BIBforeignlanguage{eng}{Medical Physics}},
  vol.~48, no.~11, pp. 7112--7126, 2021.

\bibitem{Zhang2021a}
Y.~Zhang, ``\BIBforeignlanguage{eng}{An unsupervised {2D}-{3D} deformable
  registration network ({2D3D}-{RegNet}) for cone-beam {CT} estimation.}''
  \emph{\BIBforeignlanguage{eng}{Physics in Medicine and Biology}}, vol.~66,
  no.~7, p. 074001, 2021.

\bibitem{Dweek2022}
S.~Dweek, S.~Dhou, and T.~Shanableh, ``In-{Between} {Projection}
  {Interpolation} in {Cone}-{Beam} {CT} {Imaging} using {Convolutional}
  {Neural} {Networks},'' in \emph{Medical Imaging 2022: Physics of Medical
  Imaging}, W.~Zhao and L.~Yu, Eds., vol. 12031, 2022, p. 1203129.

\bibitem{Sun2021}
L.~Sun, Z.~Jiang, Y.~Chang, and L.~Ren, ``\BIBforeignlanguage{eng}{Building a
  patient-specific model using transfer learning for four-dimensional cone beam
  computed tomography augmentation},''
  \emph{\BIBforeignlanguage{eng}{Quantitative imaging in medicine and
  surgery}}, vol.~11, no.~2, pp. 540--555, 2021.

\bibitem{Hansen2018}
D.~C. Hansen, G.~Landry, F.~Kamp, M.~Li, C.~Belka, K.~Parodi, and C.~Kurz,
  ``\BIBforeignlanguage{eng}{{ScatterNet}: {A} convolutional neural network for
  cone-beam {CT} intensity correction.}''
  \emph{\BIBforeignlanguage{eng}{Medical Physics}}, vol.~45, no.~11, pp.
  4916--4926, 2018.

\bibitem{Gottschalk2022}
T.~M. Gottschalk, A.~Maier, F.~Kordon, and B.~W. Kreher,
  ``\BIBforeignlanguage{eng}{{DL}-based inpainting for metal artifact reduction
  for cone beam {CT} using metal path length information.}''
  \emph{\BIBforeignlanguage{eng}{Medical Physics}}, vol.~50, no.~1, pp.
  128--141, 2023.

\bibitem{Maier2018}
J.~Maier, Y.~Berker, S.~Sawall, and M.~Kachelriess, ``Deep {Scatter}
  {Estimation} ({DSE}): {Feasibility} of {Using} a {Deep} {Convolutional}
  {Neural} {Network} for {Real}-{Time} {X}-{Ray} {Scatter} {Prediction} in
  {Cone}-{Beam} {CT},'' in \emph{Medical Imaging 2018: Physics of Medical
  Imaging}, J.~Lo, T.~Schmidt, and G.~Chen, Eds., vol. 10573, 2018.

\bibitem{Maier2019}
J.~Maier, E.~Eulig, T.~Vöth, M.~Knaup, J.~Kuntz, S.~Sawall, and
  M.~Kachelrieß, ``\BIBforeignlanguage{eng}{Real-time scatter estimation for
  medical {CT} using the deep scatter estimation: {Method} and robustness
  analysis with respect to different anatomies, dose levels, tube voltages, and
  data truncation.}'' \emph{\BIBforeignlanguage{eng}{Medical Physics}},
  vol.~46, no.~1, pp. 238--249, 2019.

\bibitem{Heyden2022}
B.~van~der Heyden, S.~Roden, R.~Dok, S.~Nuyts, and E.~Sterpin,
  ``\BIBforeignlanguage{eng}{Virtual monoenergetic micro-{CT} imaging in mice
  with artificial intelligence.}'' \emph{\BIBforeignlanguage{eng}{Scientific
  reports}}, vol.~12, no.~1, p. 2324, 2022.

\bibitem{Sun_2010}
M.~Sun and J.~M. Star-Lack, ``Improved scatter correction using adaptive
  scatter kernel superposition,'' \emph{Physics in Medicine and Biology},
  vol.~55, no.~22, p. 6695, 2010.

\bibitem{Nomura2019}
Y.~Nomura, Q.~Xu, H.~Shirato, S.~Shimizu, and L.~Xing,
  ``\BIBforeignlanguage{eng}{Projection-domain scatter correction for cone beam
  computed tomography using a residual convolutional neural network},''
  \emph{\BIBforeignlanguage{eng}{Medical Physics}}, vol.~46, no.~7, pp.
  3142--3155, 2019.

\bibitem{Lalonde2020}
A.~Lalonde, B.~Winey, J.~Verburg, H.~Paganetti, and G.~C. Sharp,
  ``\BIBforeignlanguage{eng}{Evaluation of {CBCT} scatter correction using deep
  convolutional neural networks for head and neck adaptive proton therapy.}''
  \emph{\BIBforeignlanguage{eng}{Physics in Medicine and Biology}}, vol.~65,
  no.~24, p. 245022, 2020.

\bibitem{Alam2021}
S.~R. Alam, T.~Li, P.~Zhang, S.-Y. Zhang, and S.~Nadeem,
  ``\BIBforeignlanguage{eng}{Generalizable cone beam {CT} esophagus
  segmentation using physics-based data augmentation.}''
  \emph{\BIBforeignlanguage{eng}{Physics in medicine and biology}}, vol.~66,
  no.~6, p. 065008, 2021.

\bibitem{liu2015faceattributes}
Z.~Liu, P.~Luo, X.~Wang, and X.~Tang, ``Deep learning face attributes in the
  wild,'' in \emph{Proceedings of International Conference on Computer Vision
  (ICCV)}, 2015, pp. 3730--3738.

\bibitem{karras2018progressive}
T.~Karras, T.~Aila, S.~Laine, and J.~Lehtinen, ``Progressive growing of {GAN}s
  for improved quality, stability, and variation,'' in \emph{International
  Conference on Learning Representations (ICLR)}, 2018.

\bibitem{Liang2019}
X.~Liang, L.~Chen, D.~Nguyen, Z.~Zhou, X.~Gu, M.~Yang, J.~Wang, and S.~Jiang,
  ``\BIBforeignlanguage{eng}{Generating synthesized computed tomography ({CT})
  from cone-beam computed tomography ({CBCT}) using {CycleGAN} for adaptive
  radiation therapy.}'' \emph{\BIBforeignlanguage{eng}{Physics in medicine and
  biology}}, vol.~64, no.~12, p. 125002, 2019.

\bibitem{Zaech2019}
J.~Zaech, C.~Gao, B.~Bier, R.~Taylor, A.~Maier, N.~Navab, and M.~Unberath,
  ``Learning to {Avoid} {Poor} {Images}: {Towards} {Task}-aware {C}-arm
  {Cone}-beam {CT} {Trajectories},'' in \emph{Medical Image Computing and
  Computer Assisted Intervention (MICCAI 2019)}, D.~Shen, T.~Liu, T.~Peters,
  L.~Staib, C.~Essert, S.~Zhou, P.~Yap, and A.~Khan, Eds., vol. 11768, 2019,
  pp. 11--19.

\bibitem{Wu2020a}
P.~Wu, N.~Sheth, A.~Sisniega, A.~Uneri, R.~Han, R.~Vijayan, P.~Vagdargi,
  B.~Kreher, H.~Kunze, G.~Kleinszig, S.~Vogt, S.~Lo, N.~Theodore, and
  J.~Siewerdsen, ``Method for {Metal} {Artifact} {Avoidance} in {C}-{Arm}
  {Cone}-{Beam} {CT},'' in \emph{Medical Imaging 2020: Physics of Medical
  Imaging}, G.~Chen and H.~Bosmans, Eds., vol. 11312, 2020.

\bibitem{Wu2020}
P.~Wu, N.~Sheth, A.~Sisniega, A.~Uneri, R.~Han, R.~Vijayan, P.~Vagdargi,
  B.~Kreher, H.~Kunze, G.~Kleinszig, S.~Vogt, S.~F. Lo, N.~Theodore, and J.~H.
  Siewerdsen, ``\BIBforeignlanguage{eng}{C-arm orbits for metal artifact
  avoidance ({MAA}) in cone-beam {CT}.}''
  \emph{\BIBforeignlanguage{eng}{Physics in medicine and biology}}, vol.~65,
  no.~16, p. 165012, 2020.

\bibitem{Uneri2021}
A.~Uneri, P.~Wu, C.~K. Jones, P.~Vagdargi, R.~Han, P.~A. Helm, M.~G. Luciano,
  W.~S. Anderson, and J.~H. Siewerdsen, ``\BIBforeignlanguage{eng}{Deformable
  {3D}-{2D} registration for high-precision guidance and verification of
  neuroelectrode placement},'' \emph{\BIBforeignlanguage{eng}{Physics in
  medicine and biology}}, vol.~66, no.~21, p. 215014, 2021.

\bibitem{Wang2022}
X.~Wang, W.~Jian, B.~Zhang, L.~Zhu, Q.~He, H.~Jin, G.~Yang, C.~Cai, H.~Meng,
  X.~Tan, F.~Li, and Z.~Dai, ``Synthetic {CT} generation from cone-beam {CT}
  using deep-learning for breast adaptive radiotherapy,'' \emph{Journal of
  Radiation Research and Applied Sciences}, vol.~15, no.~1, pp. 275--282, 2022.

\bibitem{isola2017image}
P.~Isola, J.-Y. Zhu, T.~Zhou, and A.~A. Efros, ``Image-to-image translation
  with conditional adversarial networks,'' in \emph{Proceedings of the {IEEE}
  conference on computer vision and pattern recognition (CVPR)}, 2017, pp.
  5967--5976.

\bibitem{Fan2022}
F.~Fan, B.~Kreher, H.~Keil, A.~Maier, and Y.~Huang,
  ``\BIBforeignlanguage{eng}{Fiducial marker recovery and detection from
  severely truncated data in navigation-assisted spine surgery.}''
  \emph{\BIBforeignlanguage{eng}{Medical physics}}, vol.~49, no.~5, pp.
  2914--2930, 2022.

\bibitem{Cao2022}
Z.~Cao, X.~Gao, Y.~Chang, G.~Liu, and Y.~Pei, ``\BIBforeignlanguage{eng}{A
  novel approach for eliminating metal artifacts based on {MVCBCT} and
  {CycleGAN}},'' \emph{\BIBforeignlanguage{eng}{Frontiers in Oncology}},
  vol.~12, p. 1024160, 2022.

\bibitem{Beaudry2019}
J.~Beaudry, P.~Esquinas, and C.~Shieh, ``Learning from our neighbours: a novel
  approach on sinogram completion using bin-sharing and deep learning to
  reconstruct high quality {4DCBCT},'' in \emph{Medical Imaging 2019: Physics
  of Medical Imaging}, T.~Schmidt, G.~Chen, and H.~Bosmans, Eds., vol. 10948,
  2019.

\bibitem{Lee2019a}
D.~Lee, K.~Kim, W.~Kim, S.~Kang, C.~Park, H.~Cho, Y.~Lim, G.~Kim, S.~Park,
  H.~Lim, H.~Lee, D.~Jeon, J.~Park, C.~Seo, and M.~Lee, ``Four-{Dimensional}
  {CBCT} {Reconstruction} {Based} on a {Residual} {Convolutional} {Neural}
  {Network} for {Improving} {Image} {Quality},'' \emph{Journal of the Korean
  Physical Society}, vol.~75, no.~1, pp. 73--79, 2019.

\bibitem{Zhi2019}
S.~Zhi, J.~Duan, J.~Cai, and X.~Mou, ``Artifacts {Reduction} {Method} for
  {Phase}-resolved {Cone}-{Beam} {CT} ({CBCT}) {Images} via a {Prior}-{Guided}
  {CNN},'' in \emph{Medical Imaging 2019: Physics of Medical Imaging},
  T.~Schmidt, G.~Chen, and H.~Bosmans, Eds., vol. 10948, 2019.

\bibitem{Madesta2020}
F.~Madesta, T.~Sentker, T.~Gauer, and R.~Werner,
  ``\BIBforeignlanguage{eng}{Self-contained deep learning-based boosting of
  {4D} cone-beam {CT} reconstruction},'' \emph{\BIBforeignlanguage{eng}{Medical
  physics}}, vol.~47, no.~11, pp. 5619--5631, 2020.

\bibitem{Zhi2021}
{S. Zhi}, {M. Kachelrieß}, {F. Pan}, and {X. Mou}, ``{CycN}-{Net}: {A}
  {Convolutional} {Neural} {Network} {Specialized} for {4D} {CBCT} {Images}
  {Refinement},'' \emph{IEEE Transactions on Medical Imaging}, vol.~40, no.~11,
  pp. 3054--3064, 2021.

\bibitem{Zhi2021a}
S.~Zhi and X.~Mou, ``{tN}-net: {A} {Spatiotemporal} plus {Prior} {Image}-based
  {Convolutional} {Neural} {Network} for {4D}-{CBCT} {Reconstructions}
  {Enhancement},'' in \emph{Medical Imaging 2021: Physics of Medical Imaging},
  H.~Bosmans, W.~Zhao, and L.~Yu, Eds., vol. 11595, 2021.

\bibitem{Zhang2022}
Z.~Zhang, M.~Huang, Z.~Jiang, Y.~Chang, K.~Lu, F.-F. Yin, P.~Tran, D.~Wu,
  C.~Beltran, and L.~Ren, ``\BIBforeignlanguage{eng}{Patient-specific deep
  learning model to enhance {4D}-{CBCT} image for radiomics analysis.}''
  \emph{\BIBforeignlanguage{eng}{Physics in medicine and biology}}, vol.~67,
  no.~8, p. 085003, 2022.

\bibitem{Zhang2022c}
Z.~Zhang, J.~Liu, D.~Yang, U.~S. Kamilov, and G.~D. Hugo,
  ``\BIBforeignlanguage{eng}{Deep learning-based motion compensation for
  four-dimensional cone-beam computed tomography ({4D}-{CBCT})
  reconstruction.}'' \emph{\BIBforeignlanguage{eng}{Medical physics}}, vol.~50,
  no.~2, pp. 808--820, 2023.

\bibitem{Madesta2019}
F.~Madesta, T.~Gauer, T.~Sentker, and R.~Werner, ``Self-consistent deep
  learning-based boosting of {4D} cone-beam computed tomography
  reconstruction,'' in \emph{Medical Imaging 2019: Image Processing},
  E.~Angelini and B.~Landman, Eds., vol. 10949, 2019.

\bibitem{Thummerer2022}
A.~Thummerer, C.~Seller~Oria, P.~Zaffino, S.~Visser, A.~Meijers,
  G.~Guterres~Marmitt, R.~Wijsman, J.~Seco, J.~A. Langendijk, A.~C. Knopf,
  M.~F. Spadea, and S.~Both, ``\BIBforeignlanguage{eng}{Deep learning-based
  {4D}-synthetic {CTs} from sparse-view {CBCTs} for dose calculations in
  adaptive proton therapy.}'' \emph{\BIBforeignlanguage{eng}{Medical physics}},
  vol.~49, no.~11, pp. 6824--6839, 2022.

\bibitem{shieh2018aapm}
\BIBentryALTinterwordspacing
C.~Shieh, X.~Jia, B.~Li, Y.~Gonzalez, S.~Rit, and P.~Keall, ``{AAPM grand
  challenge: SPARE—sparse-view reconstruction challenge for 4D cone-beam
  CT},'' in \emph{American Association of Physicists in Medicine Annual Meeting
  2018}, 2018. [Online]. Available:
  \url{http://amos3.aapm.org/abstracts/pdf/137-41908-452581-138082-560681006.pdf}
\BIBentrySTDinterwordspacing

\bibitem{shieh2019spare}
C.-C. Shieh, Y.~Gonzalez, B.~Li, X.~Jia, S.~Rit, C.~Mory, M.~Riblett, G.~Hugo,
  Y.~Zhang, Z.~Jiang \emph{et~al.}, ``{SPARE: Sparse-view reconstruction
  challenge for 4D cone-beam CT from a 1-min scan},'' \emph{Medical physics},
  vol.~46, no.~9, pp. 3799--3811, 2019.

\bibitem{Jiang2022a}
{Z. Jiang}, {Z. Zhang}, {Y. Chang}, {Y. Ge}, {F. -F. Yin}, and {L. Ren},
  ``Enhancement of 4-{D} {Cone}-{Beam} {Computed} {Tomography} ({4D}-{CBCT})
  {Using} a {Dual}-{Encoder} {Convolutional} {Neural} {Network} ({DeCNN}),''
  \emph{IEEE Transactions on Radiation and Plasma Medical Sciences}, vol.~6,
  no.~2, pp. 222--230, 2022.

\bibitem{Liu2019}
{J. Liu}, {Y. Kang}, {D. Hu}, and {Y. Chen}, ``{4D}-{CBCT} {Reconstruction} via
  {Motion} {Compensataion} {Learning} {Induced} {Sparse} {Tensor}
  {Constraint},'' in \emph{12th {Intl.} {Congress} on {Image} and {Signal}
  {Processing}, {BioMedical} {Engineering} and {Informatics} ({CISP}-{BMEI})},
  2019, pp. 1--5.

\bibitem{Zhang2019}
Y.~Zhang, X.~Huang, and J.~Wang, ``\BIBforeignlanguage{eng}{Advanced
  4-dimensional cone-beam computed tomography reconstruction by combining
  motion estimation, motion-compensated reconstruction, biomechanical modeling
  and deep learning.}'' \emph{\BIBforeignlanguage{eng}{Visual computing for
  industry, biomedicine, and art}}, vol.~2, no.~1, p.~23, 2019.

\bibitem{Huang2020}
X.~Huang, Y.~Zhang, L.~Chen, and J.~Wang,
  ``\BIBforeignlanguage{eng}{U-net-based deformation vector field estimation
  for motion-compensated {4D}-{CBCT} reconstruction.}''
  \emph{\BIBforeignlanguage{eng}{Medical physics}}, vol.~47, no.~7, pp.
  3000--3012, 2020.

\bibitem{Sisniega2020}
A.~Sisniega, S.~Capostagno, W.~Zbijewski, J.~Stayman, C.~Weiss, T.~Ehtiati, and
  J.~Siewerdsen, ``Estimation of {Local} {Deformable} {Motion} in
  {Image}-{Based} {Motion} {Compensation} for {Interventional} {Cone}-{Beam}
  {CT},'' in \emph{Medical Imaging 2020: Physics of Medical Imaging}, G.~Chen
  and H.~Bosmans, Eds., vol. 11312, 2020.

\bibitem{Sisniega2021}
A.~Sisniega, H.~Huang, W.~Zbijewski, J.~Stayman, C.~Weiss, T.~Ehtiati, and
  J.~Siewerdsen, ``Deformable {Image}-{Based} {Motion} {Compensation} for
  {Interventional} {Cone}-{Beam} {CT} with a {Learned} {Autofocus} {Metric},''
  in \emph{Medical Imaging 2021: Physics of Medical Imaging}, H.~Bosmans,
  W.~Zhao, and L.~Yu, Eds., vol. 11595, 2021.

\bibitem{huang2022}
H.~Huang, J.~H. Siewerdsen, W.~Zbijewski, C.~R. Weiss, M.~Unberath, and
  A.~Sisniega, ``Context-aware, reference-free local motion metric for cbct
  deformable motion compensation,'' in \emph{7th International Conference on
  Image Formation in X-Ray Computed Tomography}, J.~W. Stayman, Ed., vol.
  12304, 2022.

\bibitem{Jiang2022b}
Z.~Jiang, Y.~Chang, Z.~Zhang, F.-F. Yin, and L.~Ren,
  ``\BIBforeignlanguage{eng}{Fast four-dimensional cone-beam computed
  tomography reconstruction using deformable convolutional networks.}''
  \emph{\BIBforeignlanguage{eng}{Medical physics}}, vol.~49, no.~10, pp.
  6461--6476, 2022.

\bibitem{herzig2022deep}
I.~Herzig, P.~Paysan, S.~Scheib, A.~Z{\"u}st, F.-P. Schilling, J.~Montoya,
  M.~Amirian, T.~Stadelmann, P.~Eggenberger~Hotz, R.~M. F{\"u}chslin
  \emph{et~al.}, ``{Deep learning-based simultaneous multi-phase deformable
  image registration of sparse 4D-CBCT},'' \emph{Medical Physics}, vol.~49,
  no.~6, pp. e325--e326, 2022.

\bibitem{Hansch2019}
A.~Hansch, V.~Dicken, J.~Klein, T.~Morgasb, B.~Haas, and H.~Hahn,
  ``Artifact-driven sampling schemes for robust female pelvis {CBCT}
  segmentation using deep learning,'' in \emph{Medical Imaging 2019:
  Computer-Aided Diagnosis}, K.~Mori and H.~Hahn, Eds., vol. 10950, 2019.

\bibitem{xia2023sub}
\BIBentryALTinterwordspacing
W.~Xia, C.~Niu, W.~Cong, and G.~Wang, ``Sub-volume-based denoising diffusion
  probabilistic model for cone-beam ct reconstruction from incomplete data,''
  \emph{Computing Research Repository ({CoRR})}, vol. abs/2303.12861, 2023.
  [Online]. Available: \url{\url{https://arxiv.org/abs/2303.12861}}
\BIBentrySTDinterwordspacing

\bibitem{Wang2021}
Z.~Wang, C.~Vandersteen, T.~Demarcy, D.~Gnansia, C.~Raffaelli, N.~Guevara, and
  H.~Delingette, ``\BIBforeignlanguage{eng}{Inner-ear augmented metal artifact
  reduction with simulation-based {3D} generative adversarial networks.}''
  \emph{\BIBforeignlanguage{eng}{Computerized medical imaging and graphics :
  the official journal of the Computerized Medical Imaging Society}}, vol.~93,
  p. 101990, Oct. 2021, place: United States.

\bibitem{Imae2020}
T.~Imae, S.~Kaji, S.~Kida, K.~Matsuda, S.~Takenaka, A.~Aoki, T.~Nakamoto,
  S.~Ozaki, K.~Nawa, H.~Yamashita, K.~Nakagawa, and O.~Abe,
  ``\BIBforeignlanguage{jpn}{{Improvement} in {Image} {Quality} of {CBCT}
  during {Treatment} by {Cycle} {Generative} {Adversarial} {Network}},''
  \emph{\BIBforeignlanguage{jpn}{Nihon Hoshasen Gijutsu Gakkai zasshi}},
  vol.~76, no.~11, pp. 1173--1184, 2020.

\bibitem{song2022solving}
\BIBentryALTinterwordspacing
Y.~Song, L.~Shen, L.~Xing, and S.~Ermon, ``Solving inverse problems in medical
  imaging with score-based generative models,'' \emph{CoRR}, vol.
  abs/2111.08005, 2022. [Online]. Available:
  \url{https://arxiv.org/abs/2111.08005}
\BIBentrySTDinterwordspacing

\bibitem{wu_wavelet-improved_2023}
W.~Wu, Y.~Wang, Q.~Liu, G.~Wang, and J.~Zhang,
  ``\BIBforeignlanguage{eng}{Wavelet-improved {Score}-based {Generative}
  {Model} for {Medical} {Imaging}},'' \emph{\BIBforeignlanguage{eng}{IEEE
  transactions on medical imaging}}, vol.~PP, Oct. 2023.

\bibitem{ddpm}
\BIBentryALTinterwordspacing
J.~Ho, A.~Jain, and P.~Abbeel, ``{Denoising Diffusion Probabilistic Models},''
  \emph{CoRR}, vol. abs/2006.11239, 2020. [Online]. Available:
  \url{https://arxiv.org/abs/2006.11239}
\BIBentrySTDinterwordspacing

\bibitem{kawar2022denoising}
\BIBentryALTinterwordspacing
B.~Kawar, M.~Elad, S.~Ermon, and J.~Song, ``{Denoising Diffusion Restoration
  Models},'' \emph{CoRR}, vol. abs/2201.11793, 2022. [Online]. Available:
  \url{https://arxiv.org/abs/2201.11793}
\BIBentrySTDinterwordspacing

\bibitem{Dai2019}
{X. Dai}, {J. Bai}, {T. Liu}, and {L. Xie}, ``Limited-{View} {Cone}-{Beam} {CT}
  {Reconstruction} {Based} on an {Adversarial} {Autoencoder} {Network} {With}
  {Joint} {Loss},'' \emph{IEEE Access}, vol.~7, pp. 7104--7116, 2019.

\bibitem{zhou_unet_2018}
\BIBentryALTinterwordspacing
Z.~Zhou, M.~M.~R. Siddiquee, N.~Tajbakhsh, and J.~Liang, ``{UNet}++: {A}
  {Nested} {U}-{Net} {Architecture} for {Medical} {Image} {Segmentation},''
  \emph{Deep Learning in Medical Image Analysis and Multimodal Learning for
  Clinical Decision Support : 4th International Workshop, DLMIA 2018, and 8th
  International Workshop, ML-CDS 2018, held in conjunction with MICCAI 2018,
  Granada, Spain, S}, vol. 11045, pp. 3--11, Sep. 2018. [Online]. Available:
  \url{https://www.ncbi.nlm.nih.gov/pmc/articles/PMC7329239/}
\BIBentrySTDinterwordspacing

\bibitem{segars20104d}
\BIBentryALTinterwordspacing
W.~P. Segars, G.~Sturgeon, S.~Mendonca, J.~Grimes, and B.~M. Tsui, ``4d xcat
  phantom for multimodality imaging research,'' \emph{Medical physics},
  vol.~37, no.~9, pp. 4902--4915, 2010. [Online]. Available:
  \url{https://doi.org/10.1118/1.3480985}
\BIBentrySTDinterwordspacing

\bibitem{pinn}
\BIBentryALTinterwordspacing
M.~Raissi, P.~Perdikaris, and G.~E. Karniadakis, ``{Physics Informed Deep
  Learning (Part I): Data-driven Solutions of Nonlinear Partial Differential
  Equations},'' \emph{CoRR}, vol. abs/1711.10561, 2017. [Online]. Available:
  \url{http://arxiv.org/abs/1711.10561}
\BIBentrySTDinterwordspacing

\bibitem{perez-garcia_torchio_2021}
\BIBentryALTinterwordspacing
F.~P{\'e}rez-Garc{\'i}a, R.~Sparks, and S.~Ourselin, ``Torchio: a python
  library for efficient loading, preprocessing, augmentation and patch-based
  sampling of medical images in deep learning,'' \emph{Computer Methods and
  Programs in Biomedicine}, p. 106236, 2021. [Online]. Available:
  \url{https://doi.org/10.1016/j.cmpb.2021.106236}
\BIBentrySTDinterwordspacing

\bibitem{cardoso2022monai}
\BIBentryALTinterwordspacing
M.~J. Cardoso, W.~Li, R.~Brown, N.~Ma, E.~Kerfoot, Y.~Wang, B.~Murrey,
  A.~Myronenko, C.~Zhao, D.~Yang \emph{et~al.}, ``Monai: An open-source
  framework for deep learning in healthcare,'' \emph{arXiv preprint
  arXiv:2211.02701}, 2022. [Online]. Available:
  \url{https://doi.org/10.48550/arXiv.2211.02701}
\BIBentrySTDinterwordspacing

\bibitem{adrian_thummerer_2023_7781049}
\BIBentryALTinterwordspacing
A.~Thummerer, E.~Huijben, M.~Terpstra, O.~Gurney-Champion, M.~Afonso, S.~Pai,
  P.~Koopmans, M.~van Eijnatten, Z.~Perko, and M.~Maspero, ``{SynthRAD2023
  Challenge design: Synthesizing computed tomography for radiotherapy},'' 2023.
  [Online]. Available: \url{https://doi.org/10.5281/zenodo.7781049}
\BIBentrySTDinterwordspacing

\end{thebibliography}

%%%%%%%%%%%%%%%%%%%%%%%%%%%%%%%%%%%%%%%%%%%%%%%%%%%%%%%%%%%%%%%%%%%%%%%%%%%%%%%%%%%%%%%%%%%%%%%%%%%%%%%%%%

\vfill
\pagebreak

\begin{IEEEbiography}[{\includegraphics[width=1in,height=1.25in,clip,keepaspectratio]{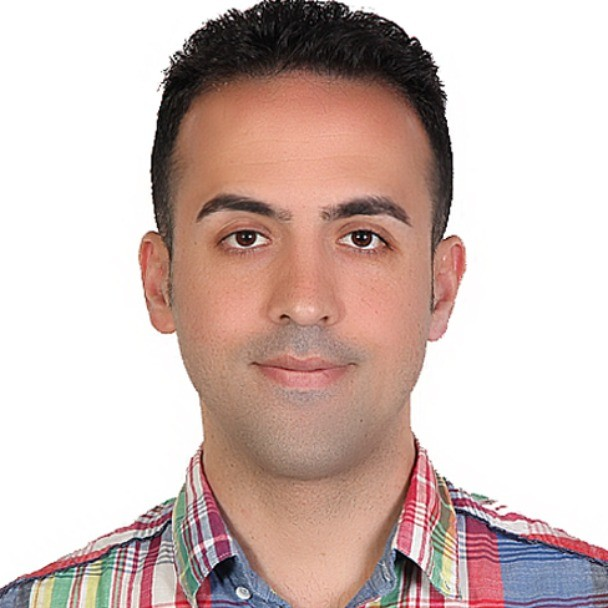}}]{Mohammadreza Amirian} 
 	received his M.Sc. degree in electrical communications technology in 2017 from Ulm University, Germany. He immediately began his Ph.D. in computer science at the Neural Information Processing Institute of Ulm University following the completion of his master's degree. During his Ph.D., he worked as a researcher at both the Institute of Applied Information Technology (InIT) and the Center for Artificial Intelligence (CAI) at Zurich University of Applied Sciences (ZHAW) in Winterthur, Switzerland. His research interests include biophysiological signal processing for person-centered medical and affective pattern recognition. Furthermore, his research pursuits extend to interpretable deep learning algorithms for medical image processing and quality enhancement in imaging technologies.
 \end{IEEEbiography}

\begin{IEEEbiography}[{\includegraphics[width=1in,height=1.25in,clip,keepaspectratio]{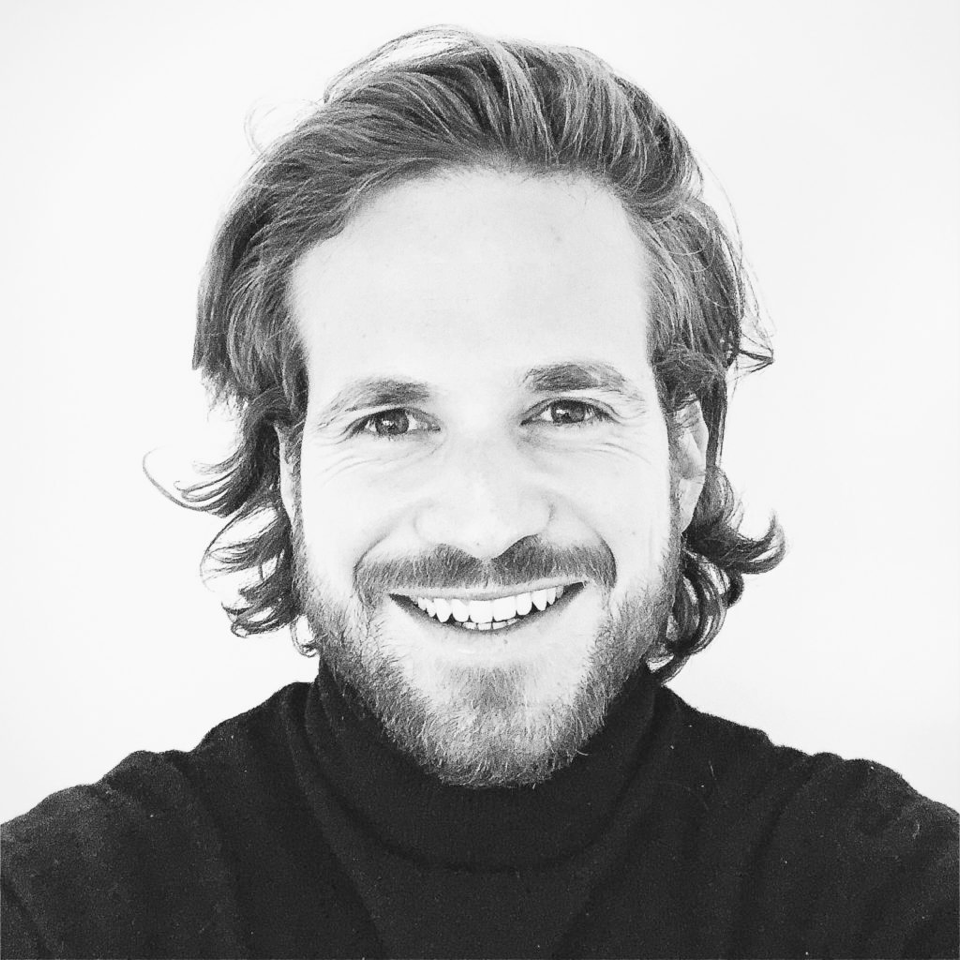}}]{Daniel Barco}
	received a M.Sc. degree in Applied Information and Data Science from the Lucerne University of Applied Sciences and Arts, in 2020. Currently, he is dedicated to advancing AI as a Ph.D. candidate at the University of Zurich (UZH) and works as a researcher at the Centre for Artificial Intelligence at the Zurich University of Applied Sciences (ZHAW), Winterthur, Switzerland. His research pursuits revolve around pioneering novel neural architectures for computer vision, while also contributing to the development of Robust and Trustworthy AI solutions.
\end{IEEEbiography}

\begin{IEEEbiography}[{\includegraphics[width=1in,height=1.25in,clip,keepaspectratio]{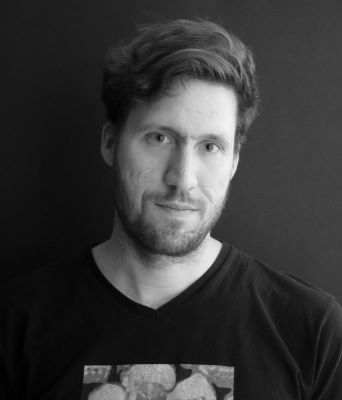}}]{Ivo Herzig} 
	is an engineer and computer scientist with professional background in software development, computational geometry, computer graphics and robotics. He is currently a Researcher at Institute of Applied Mathematics and Physics (IAMP), Zurich University of Applied Sciences (ZHAW), Switzerland, where he is focusing on deep learning for medical image analysis in the area of image-guided radiation therapy (IGRT).
\end{IEEEbiography}

\begin{IEEEbiography}[{\includegraphics[width=1in,height=1.25in,clip,keepaspectratio]{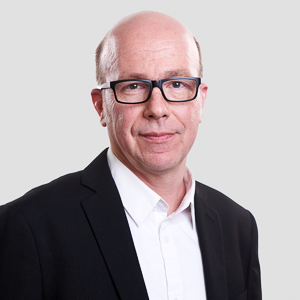}}]{Frank-Peter Schilling} 
	received his PhD degree in Physics from the University of Heidelberg, Germany in 2001. He subsequently spent many years in fundamental research at physics laboratories including CERN (Geneva, Switzerland), where he was involved in the discovery of the Higgs particle in 2012. Besides managing international scientific projects and teams, and being a top-cited author of particle physics research journal publications (h-index of 150), he developed a strong profile in computer science, big data, statistical modelling, and machine learning. He joined Zurich University of Applied Sciences ZHAW (Winterthur, Switzerland) in 2018 and is senior lecturer, group leader and deputy head of ZHAW's Centre for AI (CAI). His research interests include AI and Deep Learning, with a focus on Computer Vision (in particular for Medical Imaging), as well as on Machine Learning Operations (MLOps). In addition, he is interested in Trustworthy and Certifiable AI, as well as in applications of Deep Learning in the Physical Sciences. 
\end{IEEEbiography}

\vfill

\EOD

\end{document}